\pdfoutput=1
\documentclass[11pt]{article}

\usepackage[]{acl}

\usepackage{times}
\usepackage{latexsym}

\usepackage[T1]{fontenc}

\usepackage[utf8]{inputenc}

\usepackage{microtype}
\usepackage{graphicx}
\usepackage{booktabs}
\usepackage{colortbl}
\usepackage{multicol}
\usepackage{multirow}
\usepackage{float}
\usepackage{url}
\usepackage{makecell}
\usepackage[ruled,vlined]{algorithm2e}
\usepackage{algpseudocode}
\usepackage{subfigure}
\usepackage{mathtools}
\usepackage{nicematrix}
\usepackage{booktabs}
\usepackage{caption}

\definecolor{'deep1'}{HTML}{C5E6F8} 
\definecolor{'shallow1'}{HTML}{E4F3FC} 
\definecolor{'deep2'}{HTML}{E5F5B7} 
\definecolor{'shallow2'}{HTML}{F3FADF} 

\definecolor{'deep3'}{HTML}{FFE5C6} 
\definecolor{'shallow3'}{HTML}{FFF2E3} 
\definecolor{'deep4'}{HTML}{FFD3CF} 
\definecolor{'shallow4'}{HTML}{FFEAE8}
\definecolor{'deep5'}{HTML}{D2D0F3} 
\definecolor{'shallow5'}{HTML}{E8E7F9} 
\definecolor{darkGreen}{rgb}{0.2,0.5,0.2}
\definecolor{forestgreen}{RGB}{0,128,0}

\definecolor{forestgreen}{RGB}{0,128,0}

\title{
ChatLog: Carefully Evaluating the  Evolution of ChatGPT Across Time
}

\author{Shangqing Tu\thanks{~~Equal Contribution.}, Chunyang Li$^{*}$, Jifan Yu, Xiaozhi Wang, Lei Hou \and   Juanzi Li\\
 Department of Computer Science and Technology, Tsinghua University, Beijing 100084, China \\
  \texttt{\{tsq22,lichunya20,yujf21,wangxz20\}@mails.tsinghua.edu.cn} \\
  \texttt{\{houlei,lijuanzi\}@tsinghua.edu.cn}  \\
}

\begin{document}
\maketitle
\begin{abstract}
ChatGPT has achieved great success and can be considered to have acquired an infrastructural status. There are abundant works for evaluating ChatGPT on benchmarks. However, existing benchmarks encounter two challenges: (1) Disregard for periodical evaluation and (2) Lack of fine-grained features. In this paper, we construct ChatLog, an ever-updating dataset with large-scale records of diverse long-form ChatGPT responses for 21 NLP benchmarks from March, 2023 to now. We conduct a comprehensive performance evaluation to find that most capabilities of ChatGPT improve over time except for some abilities, and there exists a step-wise evolving pattern of ChatGPT.  We further analyze the inherent characteristics of ChatGPT by extracting the knowledge and linguistic features. We find some stable features that stay unchanged and apply them on the detection of ChatGPT-generated texts to improve the robustness of cross-version detection. We will continuously maintain our project at \url{https://github.com/THU-KEG/ChatLog/}.


\end{abstract}



\section{Introduction}
ChatGPT\footnote{\url{https://openai.com/blog/chatgpt/}} has achieved significant success~\cite{wei2023zero,liu2023evaluating,bubeck2023sparks}, solidifying its status as a foundational infrastructure. Consequently, there is a growing need for in-depth research and extensive applications based on ChatGPT's capabilities~\cite{laskar2023systematic}.

\begin{figure}[htbp] 
\centering 
\includegraphics[width=0.9\linewidth]{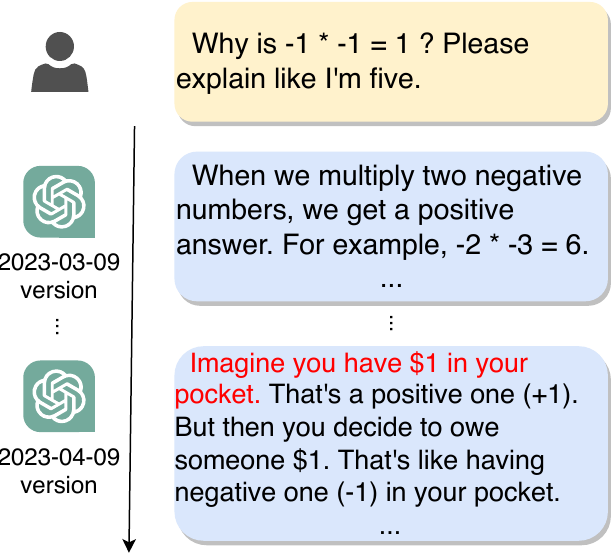} 
\caption{An example of different periods' ChatGPT answers for the same question sampled from ELI5 dataset. Later version uses a \textcolor{red}{hypothetical scenario}.}
\label{fig:motivation_example} 
\end{figure}
Whether it is to assess the capabilities of ChatGPT for building applications within a specific field or to gain insights for developing Large Language Model (LLM) from ChatGPT's performance, benchmark testing is always crucial to provide a reference. Recently, there have been a large number of newly-constructed benchmarks such as AGIEval~\cite{zhong2023agieval}, InstructEval ~\cite{chia2023instructeval} and HaluEval~\cite{HaluEval}.


However, the existing benchmarks encounter two challenges:  
(1) Lack of \textbf{periodic evaluation}: As Figure~\ref{fig:motivation_example} shows, ChatGPT is continually evolving over time. This implies benchmark results only reflect a specific period, limiting their usefulness as ChatGPT operates without a public update schedule. (2) Absence of \textbf{fine-grained features}: ChatGPT is a generative language model which inherently inclines towards producing lengthy and diverse textual outputs~\cite{guo2023close}. The linguistic characteristics exhibited in these outputs are also important part of ChatGPT's capabilities. However, existing benchmarks often tend to summarize results using a limited number of overall metrics~\cite{kocon2023chatgpt}, typically overlooking the inherent features in outputs for evaluation.

For addressing these challenges,  we have developed ChatLog, a dataset with (1) large-scale records of different ChatGPT versions that collect everyday responses from 2023-03-05 to the latest date, which is continuously updating, covering a wide range of NLP tasks~\cite{wulczyn2017ex,warstadt2019neural,rajpurkar2016squad,cobbe2021training} for automatic evaluation, and (2) long-form raw outputs of ChatGPT for fine-grained feature evaluation where each question is repeatedly tested three times to calculate the average feature, mitigating the influence of sampling randomness.


Based on the ChatLog dataset, we conduct an evaluation of ChatGPT from two perspectives: periodical evaluation and fine-grained features, leading to some interesting findings:

For periodical evaluation, we perform an automatic evaluation to observe how ChatGPT's capabilities change over time.  In tasks like text classification or sentiment analysis,  particularly in multi-label emotion analysis, the later version's ChatGPT has shown significant improvement. For example, ChatGPT's F1 score on the GoEmotions dataset~\cite{demszky2020goemotions}, which requires identifying the emotions in a piece of text, increases from $24.90$\% in June to $28.88$\% in July. Besides, ChatGPT exhibits a surprising decline in other abilities that need inference on given texts, such as NLI~\cite{wang-etal-2018-glue} and MRC~\cite{rajpurkar2016squad}, which identifies some weaknesses of ChatGPT that requires further researches. Moreover, ChatGPT exhibits a step-wise pattern where it mainly changes after one day and stays stable in the next months.  Each step change may correspond to a major update, which can help determine the effective scope of results on previous benchmarks.


For fine-grained features, we conduct feature extraction using off-the-shelf information extraction~\cite{lu-etal-2022-unified} and linguistic analysis~\cite{lee-etal-2021-pushing} tools to obtain $10$ knowledge features and $255$ linguistic features.  By calculating the variation of features, we find some features that remain consistent across time, we apply them on the ChatGPT detection task to help the detector learn more generalizable patterns. We use LightGBM~\cite{ke2017lightgbm}  to combine $10$ stable features with the RoBERTa-based~\cite{liu2019roberta} ChatGPT detector to improve $2.1$ \% accuracy of the base model on the test set consisting of generated responses from ChatGPT's new versions.  These findings yield valuable insights into features of ChatGPT and offer guidance for developers aimed at practical applications such as ChatGPT detection, ChatGPT style transfer, and LLM evaluation using ChatGPT, etc.

To summarize, our contribution is to construct the ChatLog dataset and conduct an extensive evaluation of temporal evolution and fine-grained features. ChatLog dataset will be continuously maintained to (1) facilitate the analysis of the temporal patterns in ChatGPT, providing valuable insights for assessing the valid periods of benchmark results, and (2) serve as a resource for analyzing its fine-grained features or other potential attributes.

\begin{figure}[t] 
\centering 
\includegraphics[width=\linewidth]{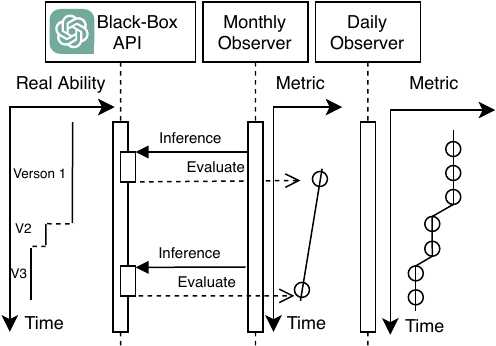} 
\caption{An illustration of periodical evaluation.}
\label{fig:define} 
\end{figure}
    
\section{ChatLog}
\label{sec:ChatLog}
To investigate how ChatGPT is changing over time, we construct ChatLog, an ever-updating dataset that records ChatGPT's responses for questions from $21$ NLU and NLG benchmarks~\cite{guo2023close,kocon2023chatgpt}. In this section, we will introduce how we collect responses periodically.

\begin{table*}[t]
    \centering
   \small
    \begin{tabular}{c| l l  |r r| c c}
        \toprule
  \textbf{Dataset} &  \textbf{Task Type} & \textbf{Task} & \textbf{\#Examples} & \textbf{Res.Len} &  \textbf{Date Range} & \textbf{\#Features} \\
        \midrule 
        
      \multirow{2}{*}{ HC3-English } & \multirow{2}{*}{ Generation } &   QA &    $25,566$ &    $172.4$ & \multirow{2}{*}{ \makecell*[c]{before \\ 2023-01-18} } & \multirow{2}{*}{$49$} \\
  &  & Dialogue & $1,337$ &   $186.4$ \\
      \midrule


         \multirow{2}{*}{ Jack\_of\_all\_trades } &   \multirow{2}{*}{ Classification } &   Pragmatic &  $18,777$ &    $2.5$ & \multirow{2}{*}{ \makecell*[c]{before \\ 2023-02-21} } & \multirow{2}{*}{0} \\
  &    &  Semantic & $19,953$ &   $7.3$ \\
        \midrule

        
      \multirow{2}{*}{ \makecell*[c]{\textbf{ChatLog-Daily} \\ \textbf{(ours)} } } & \multirow{2}{*}{ Generation } &   \multirow{2}{*}{ QA }&    \multirow{2}{*}{ $1,000 * n_d$ } &    \multirow{2}{*}{ $120.5$ } & \multirow{2}{*}{ \makecell*[c]{  2023-03-05 to \\ now  } } & \multirow{2}{*}{$265$} \\
  &  &   &  &   &   \\
      \midrule
        
    \multirow{2}{*}{\makecell*[c]{\textbf{ChatLog-Monthly} \\ \textbf{(ours)}}}
    & \multirow{2}{*}{ Classification } &   Pragmatic &   $18,777 * n_m$ &    $1.9$ &\multirow{2}{*}{ \makecell*[c]{  2023-03 to \\now} }  & \multirow{2}{*}{0} \\
  &    &  Semantic &  $19,953 * n_m$ &   $9.6$ \\
        \bottomrule
    \end{tabular}
    \caption{Statistics of each dataset. $n_d$ and $n_m$ is the number of days and months from start. \#Examples, Res.Len and \#Features is respectively the number of collected responses, average  response length, and number of extracted fine-grained features. Note that we are collecting data continuously and releasing them on GitHub periodically. }
    \label{tab:tasks_stats}
\end{table*}

\subsection{Problem  Definition}
\label{sec:problem_define}
In our study, we approach the task of tracking changes in ChatGPT as a fitting problem. As depicted in Figure~\ref{fig:define}, we make the assumption that ChatGPT's API may update over time due to many factors including the budget, the engineering constraints and the legal pressure, etc. Our work is not going to discuss why ChatGPT changes but focus on how it changes over time. Some works~\cite{chen2023chatgpt,aiyappa2023can} only use a limited number of fixed versions to study ChatGPT's changes. In contrast, we perform 
both monthly and daily evaluation to capture the nuanced changes that occur over time. 

Let $P$ represent the language modeling probability of the changing black-box model, and $t$ denote the time interval between every two evaluation with the same questions $Q$. After a time duration $T$, we can get a set of evaluation scores $\mathcal{Y}_T$ for $P$ during this period:

\begin{align}
    \mathcal{Y}_T &= \{\overline{y}(x_i)| i=1,...\frac{T}{t}\} \\
    \overline{y}(x_i)&= \frac{\sum_{1}^{N} y(x_i)}{N} \\
    x_i &= P_i (Q)
\end{align}
where $x_i$ and $P_i$ denote the responses and corresponding version's API of the $i$-th period, $y$ indicates the evaluation metric. In order to eliminate the sampling randomness of $P_i$, we repeatedly send $Q$ for $N$ times on the same  $i$-th period so that we can get the average evaluation score $\overline{y}(x_i)$ to represent $P_i$'s ability. Therefore, we can leverage each $P_i$'s score to fit the changes of $P$ over time.



\subsection{Data Collection}
\label{sec:data_collection}

\paragraph{Data Source.} To compare with previous versions' ChatGPT in open-source datasets, we select two typical datasets, including:

$\bullet$ \textbf{HC3-english}~\cite{guo2023close}. A dataset consisting of approximately $26$k questions with their corresponding human expert responses and ChatGPT responses, covering domains such as computer science, open-domain, finance, medicine, law, and psychology. 


$\bullet$ \textbf{Jack\_of\_all\_trades}~\cite{kocon2023chatgpt} A dataset to evaluate ChatGPT's performance on $25$ diverse analytical NLP tasks on existing public NLP datasets, such as sentiment analysis, emotion recognition,  and recognizing textual entailment, etc. We select $20$ of these English-based tasks and datasets that provide full evaluation scripts.

\begin{table}[t]
\scriptsize
    \centering
    \begin{tabular}{c | c}
        \toprule
  \textbf{Category}  &  \textbf{Source Dataset} \\
        \midrule

  \rowcolor{'deep1'}    & ColBERT~\cite{annamoradnejad2020colbert}\\ 
 \rowcolor{'deep1'}    & Tweeteval~\cite{barbieri-etal-2020-tweeteval} \\
  \rowcolor{'deep1'}   \multirow{-3}{*}{  \shortstack{Sentiment\\ Analysis} } & GoEmotions~\cite{demszky2020goemotions} \\
  \rowcolor{'deep2'}     & CoLa~\cite{warstadt2019neural} \\
    \rowcolor{'deep2'}     & Wikipedia Talk~\cite{wulczyn2017ex} \\
      \rowcolor{'deep2'}    & Wic~\cite{pilehvar-camacho-collados-2019-wic} \\
  \rowcolor{'deep2'}  & SMS Spam~\cite{hidalgo2012validity}\\
   \rowcolor{'deep2'}  \multirow{-5}{*}{\shortstack{Text \\ Classification}}   & Sarcasmania~\cite{Sarcasm}\\
   \rowcolor{'deep3'}    &  \\ 
   \rowcolor{'deep3'}     & \\
   \rowcolor{'deep3'}  \multirow{-3}{*}{\shortstack{Natural \\  Language \\ Inference}} & \multirow{-3}{*}{\shortstack{SuperGLUE~\cite{wang2019superglue}, \\ GLUE~\cite{wang-etal-2018-glue}}} \\

   \rowcolor{'deep4'}   & \\ 
 \rowcolor{'deep4'} & \\
 \rowcolor{'deep4'} \multirow{-3}{*}{ \shortstack{Reading \\ Comprehension} }  &  \multirow{-3}{*}{\shortstack{ SQuAD v2.0~\cite{rajpurkar2018know} \\ RACE~\cite{xu2017towards}}}  \\
   \rowcolor{'deep5'} Mathematical  &   \\ 
\rowcolor{'deep5'} Reasoning  & \multirow{-2}{*}{GSM8K~\cite{cobbe2021training}} \\
        \bottomrule
    \end{tabular}
    \caption{Source datasets of ChatLog-Monthly.}
    \label{tab:course_list}
\end{table}

\paragraph{Questions for ChatLog-Monthly.} 
As shown in Table ~\ref{tab:tasks_stats}, ChatLog-Monthly is a dataset of $38,730$ question-answer pairs collected every month. There have been 7 months' data recorded and evaluated until October, 2023. To comprehensively evaluate ChatGPT monthly, we collect all questions from Jack\_of\_all\_trades dataset which actually is composed of existing benchmarks. As shown in Table ~\ref{tab:course_list}, the questions cover a wide range of NLP tasks, including sentiment analysis~\cite{annamoradnejad2020colbert,barbieri-etal-2020-tweeteval,demszky2020goemotions},  text classification~\cite{warstadt2019neural,wulczyn2017ex,pilehvar-camacho-collados-2019-wic,hidalgo2012validity,Sarcasm}, natural language inference~\cite{wang2019superglue,wang-etal-2018-glue}, reading comprehension~\cite{rajpurkar2018know,xu2017towards} and reasoning~\cite{cobbe2021training}. Datasets' details are listed in Appendix ~\ref{appendix:benchmark_stats}.



 \begin{table*}[ht]
    
    \centering
      \small
    \begin{tabular}{l|l|c|r|r|r|r|r|r}
    \toprule
     {\makecell[c]{\textbf{Task} \\ \textbf{Name}}} & \multicolumn{1}{l|}{\makecell[c]{\textbf{Task} \\
    \textbf{Category}}} & \multicolumn{1}{r}{\makecell[c]{\textbf{ChatGPT} \\\textbf{March[\%]}}}  & \multicolumn{1}{r}{\makecell[c]{\textbf{ChatGPT}\\ \textbf{April[\%]}}} & \multicolumn{1}{r}{\makecell[c]{\textbf{ChatGPT}\\ \textbf{May[\%]}}} & \multicolumn{1}{r}{\makecell[c]{\textbf{ChatGPT}\\ \textbf{June[\%]}}} & \multicolumn{1}{r}{\makecell[c]{\textbf{ChatGPT}\\ \textbf{July[\%]}}} &\multicolumn{1}{r}{\makecell[c]{\textbf{ChatGPT}\\ \textbf{August[\%]}}} &\multicolumn{1}{r}{\makecell[c]{\textbf{ChatGPT}\\ \textbf{Sept.[\%]}}} 
    \\ \midrule
    \rowcolor{'deep1'} ColBERT & Sentiment & 74.94 & 73.00 $\sim$ & 73.30 $\sim$ & 72.40 $\sim$ & 75.24 $\sim$ & 73.01 $\sim$ & 74.84 $\sim$  \\
    \rowcolor{'shallow1'} TweetEmoji & Sentiment & 16.14 & 18.24 $\uparrow$ & 17.74 $\sim$ & 18.05 $\sim$ & 15.86 $\downarrow$ & 17.40 $\sim$ & 17.10 $\sim$ \\
    \rowcolor{'deep1'} TweetSent & Sentiment & 66.26 & 66.20 $\sim$ & 66.20 $\sim$ & 65.65 $\sim$ & 61.05 $\sim$ & 61.72 $\sim$ & 62.82 $\sim$  \\
    \rowcolor{'shallow1'} TweetStance & Sentiment & 52.02 & 54.23 $\sim$ & 52.98 $\sim$ & 53.44 $\sim$ & 48.96 $\sim$ & 51.33 $\sim$ & 51.23 $\sim$\\
    \rowcolor{'deep1'} GoEmo & Sentiment & 27.29 & 29.45 $\sim$ & 27.53 $\sim$ & 24.90 $\sim$ & 28.88 $\uparrow$ & 28.31 $\sim$ & 25.10 $\downarrow$  \\
    \rowcolor{'shallow1'} GoEmoPer0 & Sentiment & 23.97 & 23.64 $\sim$ & 24.51 $\sim$ & 21.56 $\downarrow$ & 23.81 $\uparrow$ & 22.28 $\sim$ & 22.41 $\sim$  \\
    \rowcolor{'deep1'} GoEmoPer1 & Sentiment & 22.47 & 22.32 $\sim$ & 23.85 $\sim$ & 23.09 $\sim$ & 22.68 $\sim$ & 21.62 $\sim$ & 22.87 $\sim$  \\
    \rowcolor{'shallow1'} GoEmoPer2 & Sentiment & 21.29 & 22.77 $\sim$ & 23.26 $\sim$ & 23.29 $\sim$ & 19.65 $\downarrow$ & 23.70 $\uparrow$ & 23.32 $\sim$  \\
    \rowcolor{'deep1'} GoEmoPer3 & Sentiment & 24.69 & 22.16 $\downarrow$ & 24.26 $\sim$ & 23.19 $\sim$ & 21.53 $\sim$ & 23.39 $\sim$ & 25.07 $\sim$\\
    \rowcolor{'deep2'} CoLa & Classify & 82.07 & 80.25 $\sim$ & 80.06 $\sim$ & 79.96 $\sim$ & 79.67 $\sim$ & 80.44 $\sim$ & 79.87 $\sim$  \\
    \rowcolor{'shallow2'} Aggression & Classify & 75.90 & 76.12 $\sim$ & 76.01 $\sim$ & 75.91 $\sim$ & 71.78 $\sim$ & 71.06 $\sim$ & 70.28 $\sim$  \\
    \rowcolor{'deep2'} AggressionPer & Classify & 72.97 & 72.80 $\sim$ & 72.47 $\sim$ & 72.17 $\sim$ & 69.33 $\sim$ & 69.87 $\sim$ & 69.85 $\sim$  \\
    \rowcolor{'shallow2'} WordContext & Classify & 57.99 & 62.23 $\sim$ & 62.54 $\sim$ & 61.60 $\sim$ & 62.23 $\sim$ & 59.72 $\sim$ & 57.52 $\sim$\\
    \rowcolor{'deep2'} Spam & Classify & 82.81 & 84.29 $\sim$ & 84.29 $\sim$ & 82.41 $\sim$ & 74.80 $\sim$ & 72.28 $\sim$ & 75.04 $\sim$  \\
    \rowcolor{'shallow2'} Sarcasm & Classify & 46.34 & 48.50 $\sim$ & 48.70 $\sim$ & 48.40 $\sim$ & 45.46 $\sim$ & 46.19 $\sim$ & 44.97 $\sim$ \\
    \rowcolor{'deep3'} TextEntail & NLI & 82.31 & 80.34 $\sim$ & 80.05 $\sim$ & 80.34 $\sim$ & 79.28 $\sim$ & 80.72 $\sim$ & 79.98 $\sim$  \\
    \rowcolor{'shallow3'} WNLI & NLI & 71.83 & 74.65 $\sim$ & 74.65 $\sim$ & 76.06 $\sim$ & 71.83 $\sim$ & 69.01 $\sim$ & 67.61 $\sim$  \\
    \rowcolor{'deep4'} SQuAD & MRC & 58.07 & 48.20 $\downarrow$ & 48.63 $\sim$ & 48.99 $\sim$ & 46.24 $\sim$ & 45.57 $\sim$ & 43.75 $\sim$  \\
    \rowcolor{'shallow4'} ReAding & MRC & 75.90 & 76.77 $\sim$ & 76.36 $\sim$ & 75.87 $\sim$ & 73.74 $\sim$ & 72.02 $\sim$ & 72.54 $\sim$ \\
    \rowcolor{'deep5'} GSM8K & Reasoning & 78.00 & 77.70 $\sim$ & 78.10 $\sim$ & 78.00 $\sim$ & 73.40 $\sim$ & 74.30 $\sim$ & 73.70 $\sim$  \\

    \bottomrule
    \end{tabular}
    \caption{ChatGPT's quantitative performance on each task of ChatLog-Monthly from March to September, where \emph{ChatGPT Sept.}  refers to ChatGPT's performance in September. The  $\uparrow$ or $\downarrow$ symbol refers to the performance increases or decreases over $10$\% compared with the previous month. The $\sim$  means the change is smaller than  $10$\%.}
    \label{tab:comparison_jack}
    \end{table*}
\paragraph{Questions for ChatLog-Daily.} To monitor ChatGPT's daily changes, we sample $1000$ questions from the HC3 dataset. These $1000$ questions have been repeatedly sent to ChatGPT from March 5 to now. There have been 207 days' data recorded and evaluated until the paper submission.
 The original dataset of these $1000$ questions is ELI5~\cite{fan-etal-2019-eli5}, a long-form question-answering dataset composed of threads from the Reddit forum "Explain Like I'm five". 
 The questions are open-ended and the answers are supposed to be comprehensible by a five-year-old, which means the answers should be simple and easy to understand.
  By evaluating ChatGPT on these questions, we can assess its text generation and instruction following ability. 

 \subsection{Response Process}
 \label{respones_process}
 \paragraph{OpenAI policy.}
 
 We use the API called \textit{gpt-3.5-turbo}, which is the model that powers ChatGPT and is optimized for conversational formats. According to the documentation of OpenAI API, gpt-3.5-turbo will be updated with their latest model iteration. OpenAI provides a clear pricing structure and usage policy for its API. Pricing of API calls is based on tokens. For gpt-3.5-turbo, it costs \$$0.002$ per $1000$ tokens. Furthermore, OpenAI has implemented policies to prevent abusive use of the API, including malicious requests or attempts to overload the server. As OpenAI continues to train ChatGPT on new data, we can observe how its capabilities evolve alongside the ongoing analytical progress in our research.

 \paragraph{Prompt and parameter.} 
 
 We aim to carefully address the issue of controlling variables, specifically focusing on prompts and parameters. To maintain evaluation consistency with prior research, we use the same prompt and keeping the prompts and parameters fixed. For instance, we include the prompt "explain like I'm five" after each question on ELI5~\cite{fan-etal-2019-eli5}, following the approach taken by HC3. Additionally, we utilize the default parameters provided by OpenAI during API calls. Although the default decoding parameter (temperature = $1$) introduces randomness in responses, we mitigate this by sending the same set of 1000 questions to ChatGPT three times each day. Consequently, we can calculate evaluation or feature scores by averaging the three generated results. To further ensure the stability of our results, we conduct a statistical test in Section~\ref{sec:significance_test}.

 \section{Periodical Evaluation}
 \label{sec:Evaluation}

 To investigate ChatGPT's behavior across time, we evaluate its performance on various benchmarks using automatic metrics.

 \subsection{Evaluation Settings}
 

 We quantify ChatGPT's performance on classification tasks using two classification metrics,  accuracy and macro F1 score. Accuracy measures the proportion of correctly classified instance out of the total number of instances, which provides a simple measure of classification.
 The F1 score is a commonly used metric that combines precision and recall metrics into a single score. The macro F1 score is calculated as the unweighted average of the F1 score for each class. It is used to evaluate the overall performance across all classes equally. For most tasks of ChatLog-Monthly, we use F1 as the metric except four tasks: CoLa, WordContext, WNLI and GSM8K, which utilize accuracy to measure. For evaluation on the generation task~\cite{fan-etal-2019-eli5} of ChatLog-Daily, we choose the widely-used rouge~\cite{lin2004rouge} metric to measure the token-level overlap with the reference.

 


 \begin{figure}[t]
    \centering
    \includegraphics[width=\linewidth]{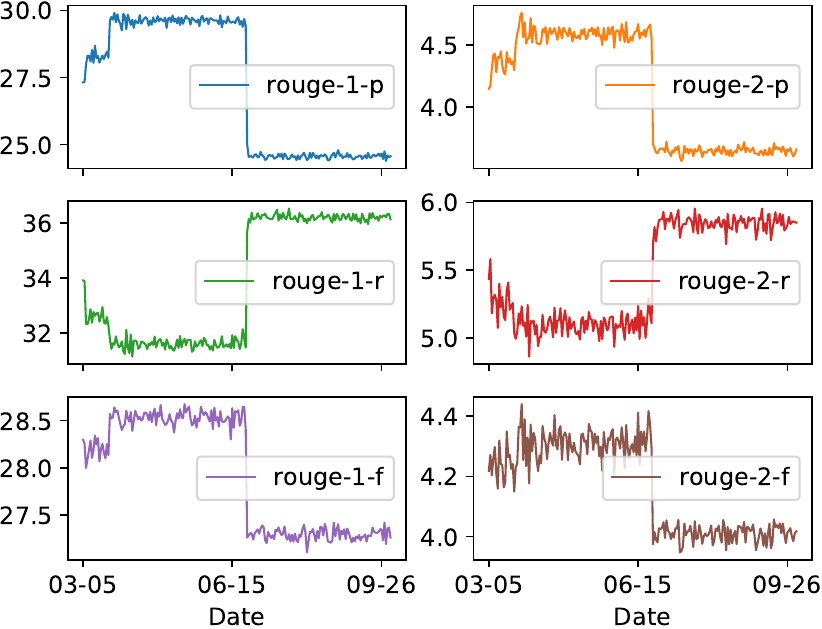}
    \caption{ 
   Trend of ChatGPT's rouge scores and linguistic features on ChatLog-Daily dataset using \textit{3 trials' average score} (i.e. $N=3$) every day since Mar 5, 2023. 
    }
    \label{fig:trend_WRich_avg_main}
\end{figure}

 \subsection{Evaluation Results}
 
 We report ChatGPT's understanding performance on ChatLog-Monthly and its generation performance on ChatLog-Daily.
 
 \label{sec:Preliminary_EXP}
 \paragraph{Understanding Performance.}
 
 In Table \ref{tab:comparison_jack}, we exhibit ChatGPT's performance after March and have the following observations:

 (1) \textbf{Progressive abilities}:  In tasks that require text classification or sentiment analysis, such as offensive language detection and multi-label emotion analysis, the later version's ChatGPT has shown significant improvement. For example, ChatGPT's F1 score on the GoEmotions dataset~\cite{demszky2020goemotions}, which requires identifying the emotions in a piece of text, increases from $24.90$\% in June to $28.88$\% in July. 
 But the performance on GoEmo drops to  $25.10$\% in September. We hypothesize that the fluctuation may be attributed to the sampling randomness as other tasks have shown stable results.

 (2) \textbf{Regressive abilities}: In tasks that rely on limited given texts, such as Natural Language Inference (NLI) and Machine Reading Comprehension (MRC), ChatGPT shows a decline in performance on these benchmarks including TextEntail,  WNLI, and SQuAD. For example, ChatGPT's F1 score on the SQuAD dataset, which requires an understanding of the given Wikipedia passage, decreased from $58.07$\% in March to $48.20$\% in April. This may be caused by the version update in March that changes the parameter of ChatGPT helping it gain some new abilities while forgetting some other aspects, leading to the drop in the MRC task.


 

 (3) \textbf{Step-wise pattern}: There are 2 notable changes are observed, one is from March to April, the other is from June to July. While minimal variations are recorded in April and May. This suggests that ChatGPT might have been updated with a new version during March and June while stay stable in April and May. These results reveal the complexities of evaluating ChatGPT as new versions may get different evaluation results over time. Our records can help determine the effective scope of evaluation results on benchmarks, satisfying the need for tracking of ChatGPT's performance.

 \paragraph{Generation Performance.}
 In addition to ChatLog-Monthly which mainly evaluates the NLU abilities, we also evaluate ChatGPT's generation ability on the ChatLog-Daily dataset with ELI5~\cite{fan-etal-2019-eli5} questions for the long-form QA task using the rouge~\cite{lin2004rouge} metric. rouge-1 and rouge-2 indicate the unigram and bigram overlap between the reference answer and the generated answer. And suffix p,r,f represents precision, recall, and f1 respectively. To rule out the possibility that the changes in ChatGPT are just from randomness and confirm the stability of our evaluation results, we conduct 3 trials for each data everyday.  We have the following findings:


 (1) As shown in Figure ~\ref{fig:trend_WRich_avg_main}, we can see that ChatGPT's generated answers are getting more concise as the rouge-1-precision score increases. While the declining performance of ChatGPT on the rouge-l-recall metric indicates that ChatGPT produces less redundant text to get higher readability.


 (2) However, only using reference-guided metrics may not evaluate the quality of generated texts comprehensively~\cite{ke2022ctrleval}. For example, the case in Figure ~\ref{fig:motivation_example} shows that ChatGPT can use the metaphor for answering questions, this semantic-level metaphoric capability can not be reflected by token-level rouge score. This limitation inspires our further evaluation in Section~\ref{sec:fg_features}.
 
 

 \begin{figure*}[htbp]
     \centering
     \includegraphics[width=\linewidth]{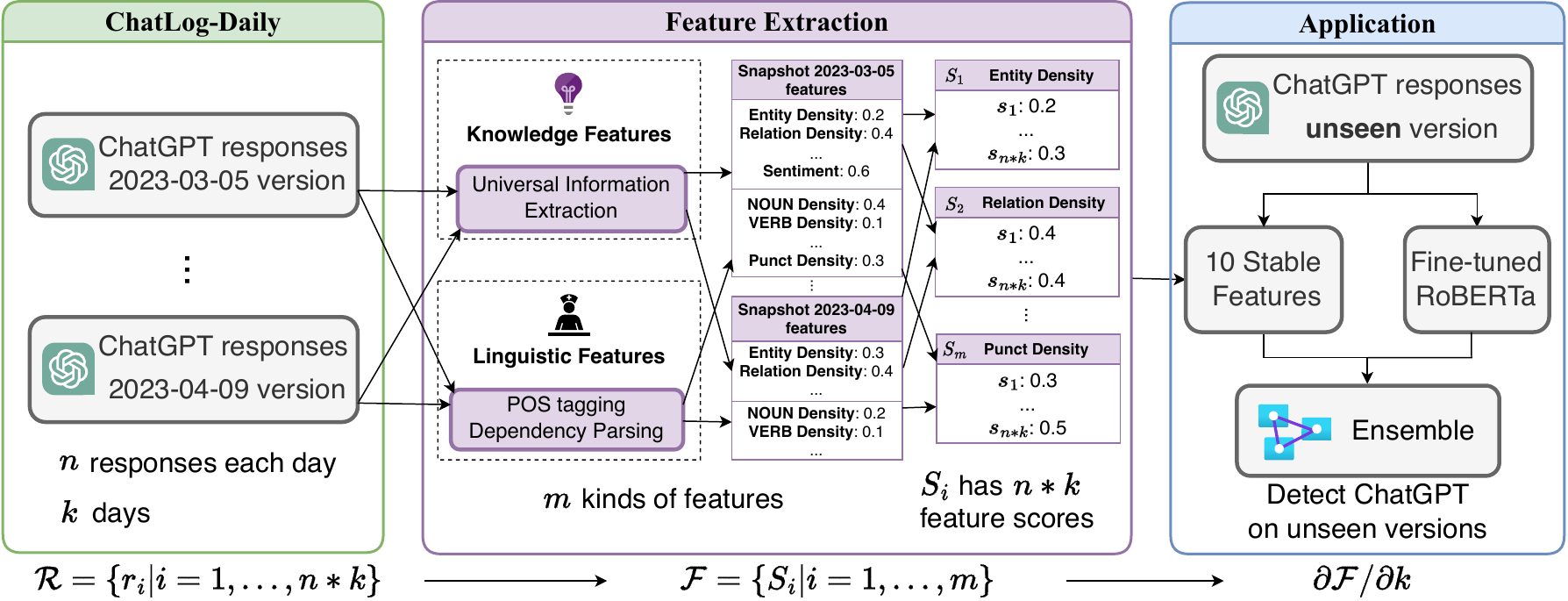}
     \caption{
     The framework of our feature extraction and application process. We first extract knowledge and linguistic features on the long-form answers from ChatLog-Daily dataset. Then we conduct an analysis on these features to find unchanged stable features over time and apply these features on the ChatGPT-generated text detection task.
     }
     \label{fig:pipeline}
 \end{figure*}

 \subsection{Statistical Test}
 \label{sec:significance_test}
 
 In addition to the visual inspection for the changes over time, we conduct a statistical test that measures the degree of variation that can be attributed to version change rather than sampling randomness. Let $T_1$ and $T_2$ denote two successive periods, if the model's version is not changed from $T_2$ to $T_1$, then their average performance $\mu_{1},\mu_{2} $ should be equal. Therefore, we can prove the significance of version change by testing the following null hypothesis,
 \begin{align}
     H_0:    \mu_{1} =  \mu_{2} 
 \end{align}
 Let $\overline{\mathcal{Y}}$ and $\mathcal{S}$  denote the average and standard deviation of evaluation scores $\mathcal{Y}$ on each period. We can calculate the $z$-score by:
 
 \begin{align}
     Z &=\frac{\overline{\mathcal{Y}_1}-\overline{\mathcal{Y}_2}}{\sqrt{\frac{\mathcal{S}_1}{|\mathcal{Y}_1|}+\frac{\mathcal{S}_2}{|\mathcal{Y}_2|}}} 
 \end{align}
 
 \begin{table}[t]
     \centering
    \small
     \begin{tabular}{c| c c c c }
         \toprule
   $T$  &   $|\mathcal{Y}|$ &   $\overline{\mathcal{Y}}$ &  $S$ & $Z$  \\
 \midrule
     03-05 to 03-23 &    $18$ &    $0.2816$ & $0.0035$ & --- \\
   03-24 to 05-09 & $46$ &    $0.2965$ &   $0.0020$ & \textbf{16.98}  \\
  05-10 to 06-27 &    $46$ &  $0.2960$ &   $0.0019$ & $-1.006$  \\
   06-28 to 08-14 &   $48$ &  $0.2457$ &   $0.0015$  & \textbf{-142.0}\\
     08-15 to 10-02 &  $49$ & $0.2457$ &   $0.0016$   & $0.1629$ \\
         \bottomrule
     \end{tabular}
     \caption{Result of the statistical test, where $Z$ is the $z$-score of the current ($T_1$) and the previous ($T_2$) period.  }
     \label{tab:z-score}
 \end{table}
 As shown in Table \ref{tab:z-score}, we calculate four different periods' $z$-score of rouge-1-p with the corresponding previous period. We find that 
 (1) There are two significant version changes, one is in March 24 with a $z$-score of $16.98$, another is in June 28 with a $z$-score of $-142$. Both of them are larger than the critical value $2.58$ of $z$-score at the $99$\% confidence level.  This is consistent with the fact that OpenAI updates their gpt-3.5-turbo API in two weeks after a version change.
 (2) Between the changes, the model's performance stays on a stable merit. For each stable version after March 24, we split their records into two evenly long parts, and the $z$-score of each part is $-1.006$ and $0.1629$ respectively, which are smaller than the critical value $1.96$ of $z$-score at the $95$\% confidence level. Therefore, we believe that ChatGPT are stable on these days.  
 

 \section{Fine-grained Features}
 \label{sec:fg_features}

 To analyze the underlying characteristics of ChatGPT beyond evaluation performance, we perform a comprehensive extraction of knowledge and linguistic features on the ChatLog-Daily dataset to discover ChatGPT generated texts' unchanged features over time. We also explore the application of these features on AI detection.
 
 \subsection{Feature Extraction Settings}
 \label{sec:task_method}
 
 Figure ~\ref{fig:pipeline} shows our pipeline and each sub-process is introduced in the following sections.
 \paragraph{Extraction Process.} We first take $n$ source queries $\mathcal{Q}$  as input then send  the query to LLM API each day across $k$ days to get $n * k$ responses $\mathcal{R}$. Then we apply off-the-shelf information and linguistic feature extraction tools~\cite{lu-etal-2022-unified,lee-etal-2021-pushing} on the raw responses $\mathcal{R}$  to extract $m$  sentiment, knowledge and  linguistic features for each $r$ in $\mathcal{R}$. Therefore, we can get a set of  $m$ snapshot feature score collections $\mathcal{F} = \{S_i| i=1,...m\}$ for the next feature analysis stage, where each snapshot $S_i$ contains $n*k$ scores.

 \begin{table}[t]
     \centering
    \small
     \begin{tabular}{l l|l r }
         \toprule
 \textbf{Type} &      \textbf{Tool}  & \textbf{Branch} & \textbf{\#Features} \\
 \midrule
      Knowledge &     UIE &  NE & $5$ \\
       Knowledge &     UIE &  OP & $2$ \\
  Knowledge &      CogIE & RE & $1$ \\
   Knowledge &      CogIE & FP & $2$ \\
     Linguistic &     LingFeat & AdSem  & $48$ \\
         Linguistic &     LingFeat & Disco  & $28$ \\
             Linguistic &     LingFeat & Synta  & $109$ \\
                 Linguistic &     LingFeat & LxSem  & $56$ \\
                     Linguistic &     LingFeat & ShaTr  & $14$ \\
         \bottomrule
     \end{tabular}
     \caption{Statistics of feature types, extractive tools, branches and the number of features under each branch.  }
     \label{tab:used_tools}
 \end{table}
 
 \paragraph{Feature Category.} As shown in Table \ref{tab:used_tools}, we use UIE~\cite{lu-etal-2022-unified} and CogIE~\cite{jin-etal-2021-cogie} tools to extract knowledge information such as named entities (NE), opinions(OP), and entity relation extraction (RE), as well as more abstract knowledge like frame-semantic parsing (FP) of semantic frames and arguments. In addition to knowledge features, we also utilize the LingFeat~\cite{lee-etal-2021-pushing} tool to obtain linguistic features such as Advanced Semantic (AdSem), Discourse (Disco), Syntactic (Synta), Lexico Semantic (LxSem) and Shallow Traditional Features (ShaTr). Overall, our extraction covers a wide range of features, with syntax and semantics being the most prominent. 

 \subsection{Detection Application Settings}
 
 We further explore the application of extracted ChatGPT features on our ChatLog-Daily dataset. We show a preliminary use case: improving the robustness of the ChatGPT detection model, which aims to classify whether a paragraph of text is generated by ChatGPT or humans. The evaluation details and baselines are as follows:

 \textbf{Dataset.} The portion of training, valid and test set is $9$:$1$:$10$. The training and the valid set consists of $1000$ ChatGPT and $1000$ human answers sampled from the ELI5 part of HC3~\cite{guo2023close} collected before January 18, 2023. For the test set, we sample $1000$ ChatGPT responses from our ChatLog-Daily dataset, which consists of ChatGPT's responses to ELI5~\cite{fan-etal-2019-eli5} questions for each day from March 5 to June 10.  We also sample 1000 human answers for the test set to make the human and ChatGPT labels balanced. 
 

 \textbf{Evaluation.} We use accuracy to evaluate the performance of 
 detection classifiers. We train models on HC3 corpus~\cite{guo2023close} with ChatGPT data gathered before January 18, 2023, and test them on our new dataset collected after March 5 to evaluate the robustness of unseen versions' data. And please refer to Appendix ~\ref{appendix:implementation_details} for further detailed implementation information.
 
 \textbf{Baselines.} Following~\citet{guo2023close}, we select 3 typical detection baselines for fine-tuning on our dataset, which are:
 
 $\bullet$ \textbf{RoBERTa}~\cite{liu2019roberta} is a pre-trained Transformer-based neural network that can be fine-tuned to handle classification tasks.
 
 $\bullet$ \textbf{PPL}~\cite{jelinek1977perplexity}  is a metric to model the probability of a word sequence on a distribution. We adapt it for the detection task.
 
 $\bullet$ \textbf{GLTR}~\cite{gehrmann-etal-2019-gltr}  utilizes the exceedingly powerful Test-2 evaluation metric, which is constructed from the number of tokens occupying the top ranks contained within the language model's predicted probability distributions. Subsequently, logistic regression techniques were applied and a model was trained to  classify input texts.



 \subsection{Feature Analysis}
 \label{sec:temporal_analysis}

\begin{table*}[t]

    \centering
   \small
    \begin{tabular}{l |c c c c c c  }
        \toprule
\multirow{2}{*}{ \textbf{Model}}  & \multirow{2}{*}{ \makecell*[c]{\textbf{ Accuracy $\uparrow$} \\ \textbf{(Unseen Dates)}}} & \multirow{2}{*}{ \makecell*[c]{\textbf{True Negative} \\ \textbf{ Rate $\uparrow$}}} & \multirow{2}{*}{ \makecell*[c]{\textbf{False Positive} \\ \textbf{ Rate $\downarrow$}}} & \multirow{2}{*}{ \makecell*[c]{\textbf{False Negative} \\ \textbf{ Rate $\downarrow$}}} & \multirow{2}{*}{ \makecell*[c]{\textbf{True Positive} \\ \textbf{ Rate $\uparrow$}}} \\ 
  &  & \\
\midrule
     RoBERTa~\cite{liu2019roberta}  & $96.2_{\pm0.1}$ & $\mathbf{100.0_{\pm0.0}}$ & $\mathbf{0.0_{\pm0.0}}$ & $7.6_{\pm0.1}$ & $92.4_{\pm0.1}$\\
      PPL~\cite{guo2023close}  & $86.7_{\pm0.5}$ & $99.2_{\pm0.1}$ & $0.8_{\pm0.1}$ & $26.0_{\pm0.4}$ & $74.0_{\pm0.4}$\\
 GLTR~\cite{gehrmann-etal-2019-gltr} & $77.6_{\pm0.4}$ & $99.2_{\pm0.1}$ & $0.6_{\pm0.1}$ & $44.2_{\pm0.4}$ & $55.8_{\pm0.4}$\\
 \midrule
RoBERTa + random 10 features  &  $95.3_{\pm0.3}$ & $98.0_{\pm1.2}$ & $1.8_{\pm1.2}$ & $7.4_{\pm3.6}$ & $92.6_{\pm3.6}$ \\
RoBERTa + stable 10 features  & $\mathbf{98.1_{\pm0.1}}$ & $98.4_{\pm0.8}$ & $1.6_{\pm0.8}$ & $\mathbf{2.2_{\pm1.2}}$ & $\mathbf{97.8_{\pm1.2}}$\\
        \bottomrule
    \end{tabular}
    \caption{ Accuracies (\%) on the test set of baselines and our detectors, along with 5 random trials' standard deviations.  }
    \label{tab:avg_HC3_classify_acc}
\end{table*}

To effectively detect ChatGPT, we conduct an analysis to discover those \textbf{stable features} that stay unchanged over time by minimizing the variation coefficient of extracted features.
As we get a set of  $m$ snapshot feature score collections $\mathcal{F} = \{S_i| i=1,...m\}$ for the analysis, where each snopshot $S_i$ contains $n*k$ scores. We can calculate the variation coefficient $c_h$ for each feature score set $S_h$:

\begin{align}
    \mu_h &=\frac{\sum_{i=1}^{n}\sum_{j=1}^{k}  s_{ij}}{n*k} \\
    \sigma_h &=\frac{\sum_{i=1}^{n}\sum_{j=1}^{k} ( s_{ij} -\frac{\sum_{l=1}^{k} s_{il}}{k})^2}{n*k} 
    \end{align}
    \begin{align}
 c_h &=\frac{\sigma_h}{\mu_h} 
\end{align}
where $h \in \{1,...,m\}$ indicates the feature's index.
Specifically, we calculate the variation coefficient on ChatLog-Daily, after filtering those features that always are zero, the rest features with minimum variation coefficient are selected. 
Finally, we select the top $10$ stable features to combine with RoBERTa for ChatGPT detection.

\subsection{Application of Stable Features}
\label{application}

\paragraph{Effectiveness of Features on ChatGPT Detection}

To demonstrate the application of the derived features, we employ them in the context of the ChatGPT detection task. While baseline detectors~\cite{guo2023close} are trained on ChatGPT corpus collected from history, the features learned from previous corpus may not stay consistent in future versions. Specifically, we use either randomly selected $10$ features or the most stable $10$ features to combine with RoBERTa. The combined $11$ features are utilized by LightGBM~\cite{ke2017lightgbm} for final label prediction.  With results in Table \ref{tab:avg_HC3_classify_acc}, we have three observations:

(1) \textbf{Performance drops on new versions}: The accuracy of RoBERTa, gltr, and ppl classifiers  decreases rapidly from the valid set with old data to the test set with newly collected corpus, which shows the difference between ordinary ChatGPT texts and up-to-date texts, and demonstrates that different periods' ChatGPT may have different text styles.  This is consistent with the fact that OpenAI updates their gpt-3.5-turbo API periodically.

(2) \textbf{Fine-tuned classifiers are accurate}: Among all classifiers, gltr and ppl actually depend on the white-box hypothesis to predict~\cite{gehrmann-etal-2019-gltr}, which have more interpretability but also encounter larger performance drop on the test dataset than the fine-tuned RoBERTa binary classifier, which achieves over $96$\% accuracy.  This is because the fine-tuned RoBERTa classifier has more parameters than gltr and ppl, which can help it to better adapt to the unseen versions.

(3) \textbf{Stable features enhance the model}: The incorporation of our 10 most stable features significantly boosts the performance of the RoBERTa detector on the test dataset, whereas the utilization of random features actually hurt the performance, which proves that our stable features can effectively improve the detector's robustness on unseen versions' ChatGPT texts. This is because the stable features are more likely to be consistent across different versions, which can help the detector to learn more generalizable patterns.

\section{Related Work}
After the success of LLMs~\cite{glaese2022improving,openai2023gpt4}, many efforts have been devoted to evaluating the capabilities of LLMs, such as natural language understanding~\cite{laskar2023systematic,das2023evaluating,HaluEval}, reasoning~\cite{frieder2023mathematical,liu2023evaluating,bian2023chatgpt}, and generation abilities~\cite{bubeck2023sparks,zhou2023recurrentgpt,gao2023enabling}. Some find that recent ChatGPT and GPT4 perform better than previous models~\cite{zhong2023agieval,chia2023instructeval}, and there may exist an evolutionary tree of language models~\cite{yang2023harnessing}. Inspired by the concept of evolution~\cite{ogundare2023comparative}, we study whether ChatGPT can evolve across its versions.

Due to the strong generation ability, LLMs have significantly impacted various fields, such as education~\cite{khalil2023will}. Students may use ChatGPT for finishing assignments and exams~\cite{haensch2023seeing,herbold2023ai}. Consequently, there is a need for effective detection methods~\cite{kirchenbauer2023watermark,mitchell2023detectgpt,krishna2023paraphrasing} to address this problem~\cite{tang2023science}. We hope our extracted features may assist LLM detection application~\cite{liu2023argugpt,li2023origin,yu2023gpt}. 
\section{Conclusion}
This study introduces ChatLog, a dynamic dataset that evaluates ChatGPT's evolution across various benchmarks. The evaluation encompasses 21 tasks, offering a comprehensive assessment of ChatGPT's evolving natural language understanding and generation capabilities over time. Additionally, we extract features from ChatGPT's daily long-form responses to investigate its intrinsic characteristics, which can significantly enhance the detection application for ChatGPT. We hope that this research will contribute to a greater understanding of ChatGPT's evolution over time and will have practical implications for real-world applications.

\section*{Limitations}

Although we have conducted extensive experiments, there are still some limitations for our work: (1) The LLM responses in our paper are all collected from ChatGPT. And all the analytical experiments are based on ChatGPT responses. Whether the observed phenomena widely exist on other LLMs is an open question and needs more explorations. (2) The datasets employed in our experiments are entirely in English, imposing limitations on the linguistic features considered in our evaluation. We chose English just because of its extensive resources, and we intuitively believe that the phenomena observed in experiments are not dependent on the English language. Therefore, we look forward to future studies that include experiments using more diverse languages. (3) The application mentioned in our paper just covers the ChatGPT detection task. The potential usages of our work are still under exploration. Besides, in the ChatGPT detection experiment, the extracted stable features are only applied on RoBERTa base model. Whether these features are useful for other models is unclear and more experiments are needed to verify it.


\section*{Ethics Statement}

In this section, we will discuss the ethical consideration for our work.
\paragraph{Licenses.} For open-accessible datasets used in our work, we have checked their licenses. The HC3~\cite{guo2023close} dataset is shared under the CC-BY-SA license and the JAT dataset~\cite{kocon2023chatgpt} is released under the MIT license.
The Licenses for the feature extraction tools are also available.  UIE~\cite{lu-etal-2022-unified} is released under the Creative Commons Attribution-NonCommercial-ShareAlike 4.0 International Public License for Noncommercial use only, CogIE~\cite{jin-etal-2021-cogie} is shared under the Apache-2.0 license, and LingFeat~\cite{lee-etal-2021-pushing} is shared under the CC-BY-SA-4.0 license,.

\paragraph{Ethics Considerations for ChatGPT} ChatGPT complies with the AI ethical guidelines set by the European Union\footnote{\url{https://digital-strategy.ec.europa.eu/en/library/ethics-guidelines-trustworthy-ai}}. These guidelines place emphasis on various ethical aspects, including human oversight, technical robustness, safety, privacy, data governance, transparency, diversity, non-discrimination, societal and environmental well-being, and accountability. They highlight the importance of empowering humans, ensuring the safety and accuracy of AI systems, promoting sustainability, and establishing accountability mechanisms for potential negative consequences. This encourages our work to record ChatGPT's daily responses that may help policy makers conduct regulations for generative AI systems like ChatGPT.
\bibliographystyle{acl_natbib}
\bibliography{custom}

\appendix


\section{Evaluation Details}
\label{appendix:benchmark_stats}

\subsection{Source Benchmark Statistics}

To compare ChatGPT's performance with its previous version and SOTA, we tested the performance of ChatGPT with the same prompts and questions in prior research~\cite{guo2023close, kocon2023chatgpt, shakarian2023independent}, the corresponding tasks or datasets are as follows: 

$\bullet$ \textbf{Aggression and AggressionPer}. The Aggression and AggressionPer tasks, which are based on the Wikipedia Talk Labels dataset~\cite{wulczyn2017ex}, are offensive language detection tasks. These tasks are presented in the form of binary classification to classify text as either aggressive or not. Specifically, AggressionPer uses the personalized variant by providing user annotations with prompts.

$\bullet$ \textbf{CoLa}. The CoLa task, which is based on The Corpus of Linguistic Acceptability~\cite{warstadt2019neural},  is a linguistic acceptability task that requires classifying input text as either grammatically correct or not.

$\bullet$ \textbf{ColBERT}. The ColBERT task, which is based on the ColBERT dataset~\cite{annamoradnejad2020colbert}, is a humor recognition task. This task is binary classification, which is to classify text as either humor or not humor.

$\bullet$ \textbf{Sarcasm} The Sarcasm task, which is based on the Sarcasmania dataset~\cite{Sarcasm}, is sarcasm classification task. This task is presented in the form of binary classification, with the task being to classify text as either sarcastic or non-sarcastic.

$\bullet$ \textbf{Spam}. The Spam task, which is based on the SMS Spam Collection v.1~\cite{hidalgo2012validity}, is a spam detection task. This task is presented in the form of classifying text as either spam or not.

$\bullet$ \textbf{WordContext}. The WordContext task, which is based on the Wic dataset~\cite{pilehvar-camacho-collados-2019-wic}, is a word sense disambiguation task. This task is in the form of binary pair classification to test whether the highlighted word in two different contexts expresses the same meaning.

$\bullet$ \textbf{TextEntail}. The TextEntail task, which is based on one of the SuperGLUE benchmark~\cite{wang2019superglue}, is recognizing textual entailment task. This task is binary sentence pair classification, with the task being to test whether the two sentences are ``entailed" or ``not\_entailed".

$\bullet$ \textbf{WNLI}. The WNLI task, which is based on the SuperGLUE Winograd NLI dataset from the GLUE benchmark~\cite{wang-etal-2018-glue}, is a natural language inference task. This task is binary sentence pair classification, which is to test whether the ambiguous pronoun in the second sentence is entailed with the pronoun in the first sentence.

$\bullet$ \textbf{SQuAD}. The SQuAD task is based on SQuAD v2.0~\cite{rajpurkar2018know}, an extractive question-answering dataset. The language model should answer the question and infer whether the answer can be found in the given context by giving its position.

$\bullet$ \textbf{GSM8K}. The GSM8K task is based on the GSM8K~\cite{cobbe2021training}, a maths word problem dataset. This task is framed as mathematical reasoning. Chain-of-thought reasoning prompting is used for this task. 

$\bullet$ \textbf{GoEmo and GoEmoPer}. The GoEmo and GoEmoPer tasks, which are based on the GoEmotions dataset~\cite{demszky2020goemotions}, are sentiment analysis tasks. These tasks are framed as multi-label classification, with the tasks being to classify the emotions in a piece of text as emotion labels in 27 possible emotions plus neutral.  Specifically, GoEmoPer tasks use individual annotator annotations with prompts.

$\bullet$ \textbf{TweetEmoji, TweetSent, TweetStance}. The TweetEmoji task, TweetSent task, and Tweet Stance task, which are based on the Tweeteval dataset~\cite{barbieri-etal-2020-tweeteval}, are emoji prediction task, sentiment analysis task and stance detection task respectively. These tasks are presented in the form of multi-class classification. The TweetEmoji task focuses on emoji prediction for a given tweet, the TweetSent task aims to classify the emotion of a given tweet as negative, neutral, or positive, and the TweetStance task focuses on detecting the stance of a piece of text as none, against or favor.

$\bullet$ \textbf{ReAding}. The ReAding task is based on the RACE dataset~\cite{xu2017towards}, a reading comprehension dataset. This task is framed as multiple choice QA to choose the most likely option of four statements.

\subsection{Evaluation Stability}

\begin{figure}[htbp]
    \centering
    \includegraphics[width=\linewidth]{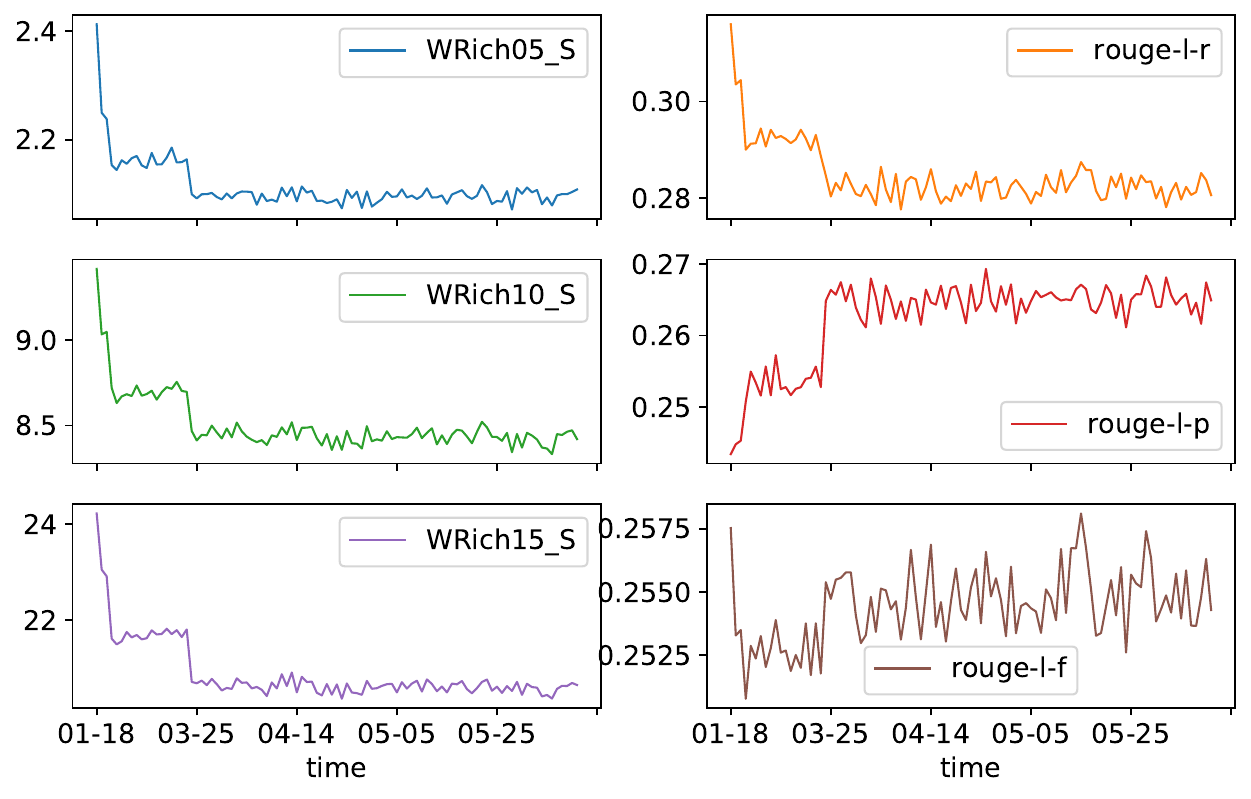}
    \caption{ 
   Trend of ChatGPT's rouge scores and linguistic features on ChatLog-Daily dataset using \textbf{only 1 trail}'s score every day from March 5 to June 10, 2023. Note that the initial data at January 18 comes from HC3~\cite{guo2023close} dataset. WRich\_S denotes Semantic Richness coverage on several topics extracted from Wikipedia.
    }
    \label{fig:trend_WRich_base1}
\end{figure}

\begin{figure}[htbp]
    \centering
    \includegraphics[width=\linewidth]{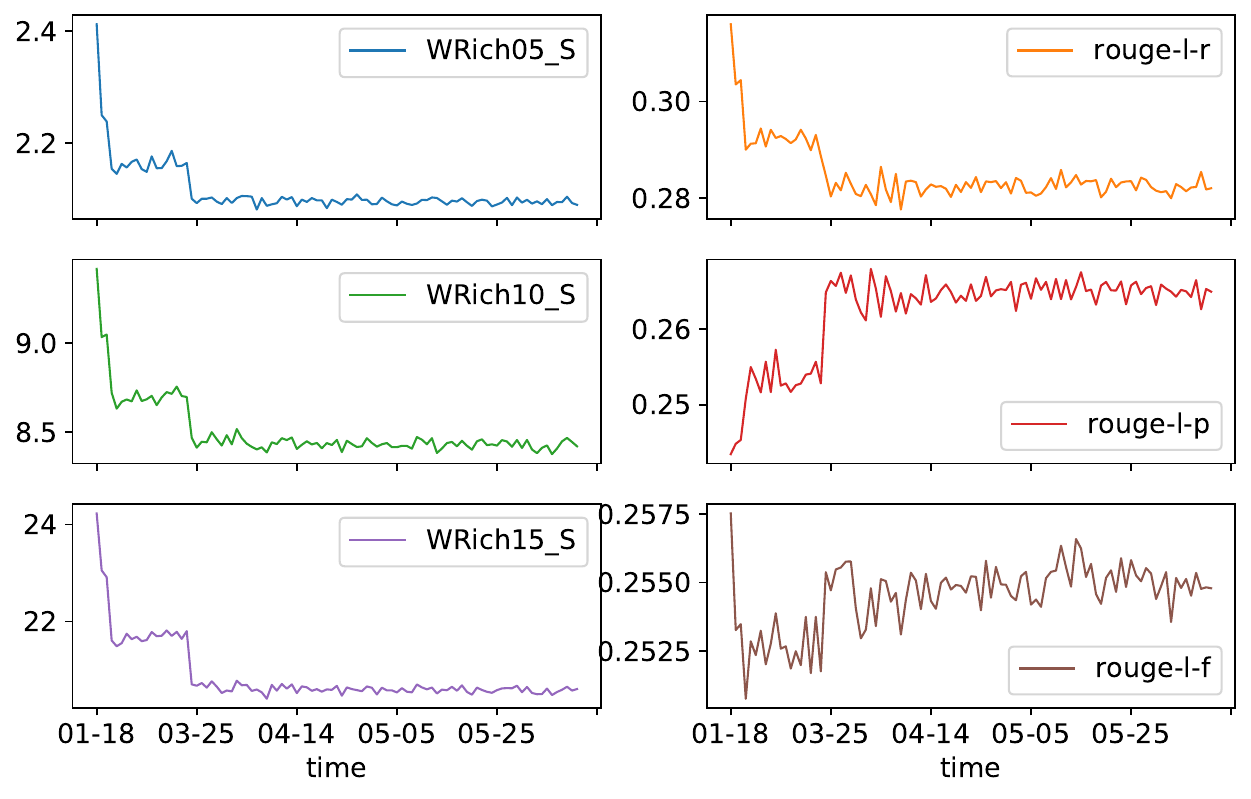}
    \caption{ 
   Trend of ChatGPT's rouge scores and linguistic features on ChatLog-Daily dataset using \textbf{3 trails' average score} every day from March 5 to June 10, 2023. Note that the initial data at January 18 comes from HC3~\cite{guo2023close} dataset. WRich\_S denotes Semantic Richness coverage on several topics extracted from Wikipedia.
    }
    \label{fig:trend_WRich_avg}
\end{figure}

To rule out the possibility that the changes in ChatGPT are just from randomness and confirm the stability of our evaluation results, we conduct 3 trails for each data everyday (sending the same question to OpenAI api for 3 times with the same hyper-parameters and prompts). We compare the difference of evaluating and analyzing ChatGPT by using results of 1 query and 3 repeated queries' average score for each data in Figure \ref{fig:trend_WRich_base1} and \ref{fig:trend_WRich_avg}. We observed that, after taking the average of three results, the fluctuations in the feature and evaluation scores of ChatGPT decrease. Specifically, in the stable range after April, the curve of only one trail's result exhibits relatively large fluctuations, but it doesn't reach the level of difference observed between March and April due to version changes. These phenomena indicate that the observed temporal variations are indeed primarily attributed to version changes rather than randomness.

\begin{figure}[htbp] 
\centering 
\includegraphics[width=\linewidth]{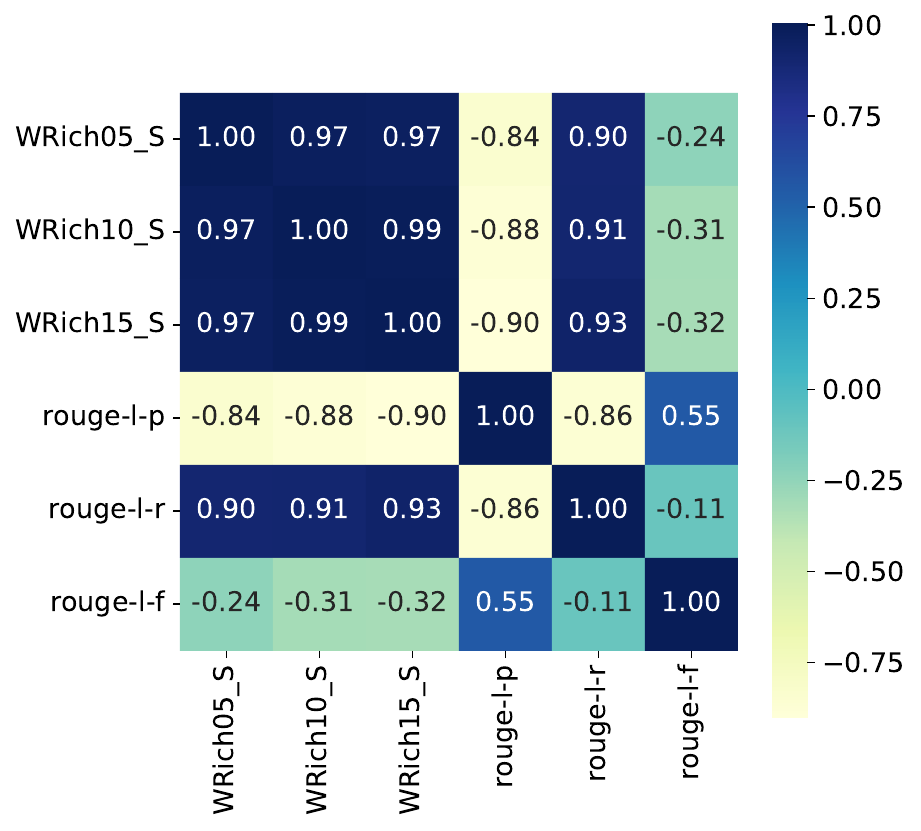} 
\caption{\textbf{Pearson correlation coefficient} between the
three rouge metrics  rouge-l-\{precision, recall, f1\} and  three semantic features  WRich\{05,10,15\}\_S, on the ChatLog-Daily dataset using \textbf{only 1 trail}'s score. WRich\_S denotes Semantic Richness coverage on several topics extracted from Wikipedia.} 
\label{fig:heat_map_base1} 
\end{figure}

\begin{figure}[htbp] 
\centering 
\includegraphics[width=\linewidth]{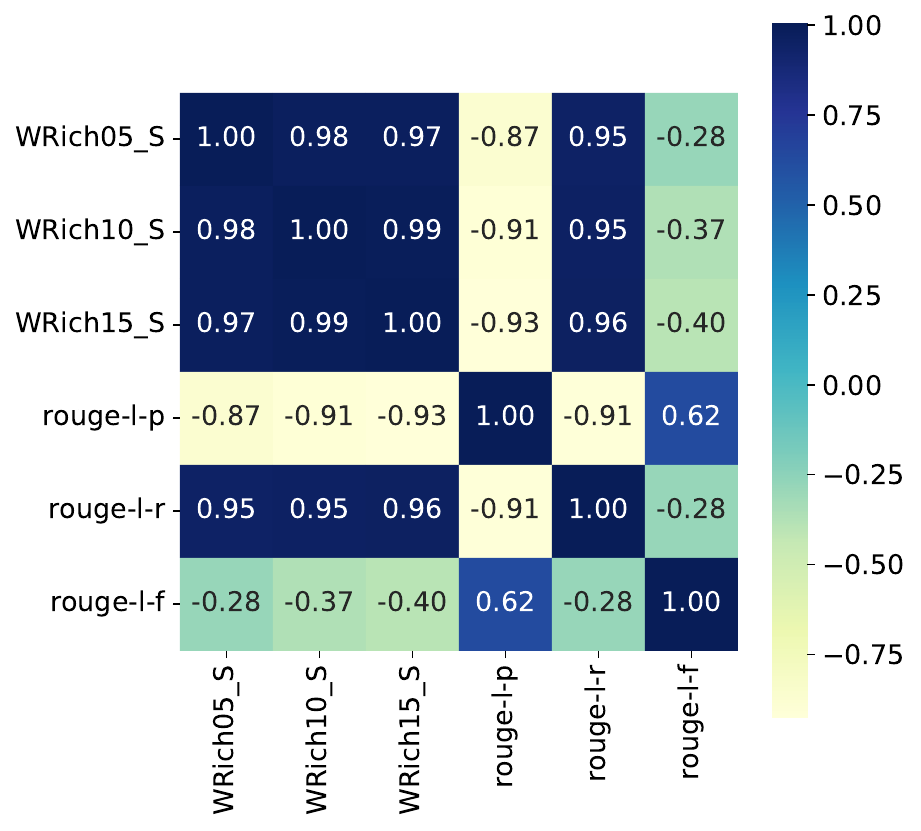} 
\caption{\textbf{Pearson correlation coefficient} between the
three rouge metrics  rouge-l-\{precision, recall, f1\} and  three semantic features  WRich\{05,10,15\}\_S, on the ChatLog-Daily dataset using \textbf{3 trails' average score}. WRich\_S denotes Semantic Richness coverage on several topics extracted from Wikipedia.} 
\label{fig:heat_map_avg} 
\end{figure} 

We also calculate the  Pearson correlation coefficient using results of 1 query and 3 repeated queries' average score for each data in Figure \ref{fig:heat_map_base1} and \ref{fig:heat_map_avg}. The results indicate a stronger correlation between the features and rouge scores of ChatGPT when using the average value of three responses. This suggests that employing the averaging approach can help ChatGPT's features regress to a reasonable range,  thus establishing a stable relationship with the rouge metric.
\subsection{Case Study}
\label{appendix:case_study}

This section contains sampled examples from the ChatLog-Monthly dataset for some evaluation tasks. ChatGPT's performance on these tasks is not perfect. When the text is short or contains little information, it may have difficulty making judgments. In addition, the prompt is not always valid, so ChatGPT may give out-of-bounds results or its output may be in a format that requires post-processing. The following examples from Table \ref{tab:appendix-aggression-correct-analysis} to Table \ref{tab:appendix-TweetEval-analysis} show some of the above situations, including absolutely correct (content and format), correct but wrong formatted, unable to answer, incorrect, out of range, etc.

\begingroup
\begin{table*}[ht]

    \centering
    \small
    \begin{tabular}{p{\linewidth}}
        \toprule

            \vspace{-2mm}
        \textbf{\textsc{Question:}} Which one of the attributes: "aggressive", "non-aggressive" describes a given text?  Text:  Lets see...no final warning, and talk-page stalking. Nice. \\
        \vspace{-1mm}
        \textbf{\textsc{Model Answer \color{blue}{(incorrect)}:}} expected: non-aggressive, answer: aggressive\\
        \midrule
        \vspace{-2mm}
        \textbf{\textsc{Question:}} Which one of the attributes: "aggressive", "non-aggressive" describes a given text? Text:  Personal Attack Check the history of that IP and you will see that it is not at all a personal attack. Please also remember that you can't blank talk pages that don't belong to you. \\
        \vspace{-1mm}
        \textbf{\textsc{Model Answer \color{blue}{(incorrect)}:}} expected: non-aggressive, answer: aggressive \\
        \midrule
        \vspace{-2mm}
        \textbf{\textsc{Question:}} Knowing that a certain person described each text with a given attribute:  Text: `Benji dude. I wouldnt bother these lot on here are clearly deluded. If MSNBC/BBC or CNN are not reliable sources because they go back to an interview then this is one of the reasons wikipedia is critizsed for, and Portia keeps reverting edits made and class them as vandalism yet coming back with ``I understand your passion``. You cannot find any retraction by Stiers or his people if you google it, if were really such an issue for Stiers he would have retracted it immediatly and it hasnt publically, sooo silly.`; Attribute: "non-aggressive" Text: Xman52 is vandalizing the Hopkins page.; Attribute: "non-aggressive" Text:  RE: the gods template The text is different, looks better then the plain ass text used on everything else. And once again, who gives a shit? I use it on a few pages, what does it matter?  ; Attribute: "non-aggressive"  write which one of the attributes: "aggressive", "non-aggressive" would this person describe a given text? Write your answer in the form of a Python list containing the appropriate attribute.  Text:  HA! That is ridiculous. I've seen his posts. He actually believes what he says and that the WTC was taken down by fire and damage. What is he a mailboy?   \\
        \vspace{-1mm}
        \textbf{\textsc{Model Answer \color{darkGreen}{(correct)}:}} expected: aggressive, answer: aggressive \\
        \bottomrule
    \end{tabular}
        \caption{
   Aggression and AggressionPer task examples.
    }
    \label{tab:appendix-aggression-correct-analysis}
\end{table*}
\endgroup

\begingroup
\begin{table*}[ht]

    \centering
    \small
    \begin{tabular}{p{\linewidth}}
        \toprule
        \vspace{-2mm}
        \textbf{\textsc{Question:}} Describe the grammatic acceptability of the given text. Choose your answer from provided list and map your answer with following unacceptable: 0, acceptable: 1 and return an integer as a result. Do not explain yourself.Input data: Text: Michelle kept the desk with the papers. Possible acceptability: unacceptable, acceptable \\
        \vspace{-1mm}
        \textbf{\textsc{Model Answer \color{blue}{(incorrect)}:}}expected: 0, answer: 1 \\
        \midrule
        \vspace{-2mm}
        \textbf{\textsc{Question:}} Describe the grammatic acceptability of the given text. Choose your answer from provided list and map your answer with following unacceptable: 0, acceptable: 1 and return an integer as a result. Do not explain yourself. Input data:  Text: Most of the fruit is rotten.  Possible acceptability: unacceptable, acceptable \\
        \vspace{-1mm}
        \textbf{\textsc{Model Answer \color{darkGreen}{(correct)}:}} expected: 1, answer: 1 \\
        \midrule
        \vspace{-2mm}
        \textbf{\textsc{Question:}} Answer should be only python list containing string either "spam" or "not spam". Content of the SMS: IMPORTANT MESSAGE. This is a final contact attempt. You have important messages waiting out our customer claims dept. Expires 13/4/04. Call 08717507382 NOW!. \\
        \vspace{-1mm}
        \textbf{\textsc{Model Answer \color{darkGreen}{(correct)}:}}expected: spam, answer: spam \\
        \midrule
        \vspace{-2mm}
        \textbf{\textsc{Question:}} Answer should be only python list containing string either "spam" or "not spam". Content of the SMS: I don't know u and u don't know me. Send CHAT to 86688 now and let's find each other! Only 150p/Msg rcvd. HG/Suite342/2Lands/Row/W1J6HL LDN. 18 years or over.. \\
        \vspace{-1mm}
        \textbf{\textsc{Model Answer \color{blue}{(incorrect)}:}} expected: spam, answer: not spam \\
        \midrule
        \vspace{-2mm}
        \textbf{\textsc{Question:}} Which one of the attributes: "funny", "not funny" describes a given text? Write your answer in the form of a Python list containing the appropriate attribute.  Text: But is it too late to try? \\
        \vspace{-1mm}
        \textbf{\textsc{Model Answer \color{blue}{(none)}:}} expected: not funny, answer: not enough information is given to determine whether the text is funny or not funny. both attributes could be possible. therefore, the python list containing the appropriate attribute cannot be provided. \\
        \midrule
        \vspace{-2mm}
        \textbf{\textsc{Question:}} Which one of the attributes: "funny", "not funny" describes a given text? Write your answer in the form of a Python list containing the appropriate attribute.  Text: Eleven days until the  National Convention! Are you as excited as we are? \#FillYourBucket \#cantwait \\
        \vspace{-1mm}
        \textbf{\textsc{Model Answer \color{darkGreen}{(correct)}:}} expected: funny, answer: ["funny"] \\
        \bottomrule
    \end{tabular}
        \caption{
   CoLa, ColBERT, Sarcasm and Spam task examples.
    }
    \label{tab:appendix-cola-correct-analysis}
\end{table*}
\endgroup

\begingroup
\begin{table*}[ht]

    \centering
    \small
    \begin{tabular}{p{\linewidth}}
        \toprule
        \vspace{-2mm}
        \textbf{\textsc{Question:}} Write the reasoning and highlight the answer at the end in format: 'Answer: '. The question is: Carlos is planting a lemon tree. The tree will cost \$90 to plant. Each year it will grow 7 lemons, which he can sell for \$1.5 each. It costs \$3 a year to water and feed the tree. How many years will it take before he starts earning money on the lemon tree?\\
        \vspace{-1mm}
        \textbf{\textsc{Model Answer \color{blue}{(incorrect and wrong understanding)}:}} expected: 13, answer: to calculate when carlos will start earning money on the lemon tree, we need to compare the income to the expenses.

each year, carlos will earn 7 lemons x \$1.5/lemon = \$10.5 in revenue. 
his yearly costs are \$3 to water/feed the tree. 

so each year, his profit will be \$10.5 - \$3 = \$7.5. 

we need to figure out how many years it will take to earn back the initial \$90 investment. 

 \$90 / \$7.5 = 12 

{\color{blue}answer: it will take 12 years before carlos starts earning money on the lemon tree.} \\

 \midrule
        \vspace{-2mm}
        \textbf{\textsc{Question:}} Which one of the attributes: "funny", "not funny" describes a given text? Write your answer in the form of a Python list containing the appropriate attribute.  Text: Eleven days until the  National Convention! Are you as excited as we are? \#FillYourBucket \#cantwait \\
        \vspace{-1mm}
        \textbf{\textsc{Model Answer \color{darkGreen}{(correct)}:}} expected: funny, answer: ["funny"] \\

         \midrule
        \vspace{-2mm}
        \textbf{\textsc{Question:}} Write the reasoning and highlight the answer at the end in format: 'Answer: '. The question is: Melanie is a door-to-door saleswoman. She sold a third of her vacuum cleaners at the green house, 2 more to the red house, and half of what was left at the orange house. If Melanie has 5 vacuum cleaners left, how many did she start with? \\
        \vspace{-1mm}
        \textbf{\textsc{Model Answer \color{blue}{(incorrect and wrong calculation)}:}} expected: 18, answer: to solve this problem, we need to work backwards from the end. melanie has 5 vacuum cleaners left after selling a third at the green house, 2 more to the red house, and half of what was left at the orange house.

let's start with the half sold at the orange house. if half of what was left after selling to the green and red houses is 5, then the total left after selling to the green and red houses must be 10.

to find out how many she started with, we need to work backwards again. we know that she sold a third at the green house and 2 more to the red house, which means that the total sold at those two houses is:

(1/3)x + 2

where x is the total number of vacuum cleaners she started with.

so we can set up an equation:

x - ((1/3)x + 2) - ((1/2)(x - (1/3)x - 2)) = 5

simplifying:

x - (1/3)x - 2 - (1/2)x + (1/6)x + 1 = 5

multiplying both sides by 6 to eliminate the fractions:

{\color{blue} 6x - 2x - 24 - 3x + x + 6 = 30}

simplifying:

2x - 18 = 30

2x = 48

x = 24

answer: melanie started with 24 vacuum cleaners. \\

         \midrule
        \vspace{-2mm}
        \textbf{\textsc{Question:}}Write the reasoning and highlight the answer at the end in format: 'Answer: '. The question is: Eliza's rate per hour for the first 40 hours she works each week is \$10. She also receives an overtime pay of 1.2 times her regular hourly rate. If Eliza worked for 45 hours this week, how much are her earnings for this week?\\
        \vspace{-1mm}
        \textbf{\textsc{Model Answer \color{darkGreen}{(correct)}:}} eliza's earnings for the first 40 hours:

40 hours x \$10/hour = \$400

eliza's overtime earnings for the additional 5 hours:

5 hours x (\$10/hour x 1.2) = \$60

total earnings for eliza this week:

\$400 + \$60 = \$460

answer: \$460\\
        \bottomrule
    \end{tabular}
        \caption{
    MathQA task examples.
    }
    \label{tab:appendix-gsm8k-correct-analysis}
\end{table*}
\endgroup

\begingroup
\begin{table*}[ht]

    \centering
    \small
    \begin{tabular}{p{\linewidth}}
        \toprule
        \vspace{-2mm}
        \textbf{\textsc{Question:}} Are the given texts expressing the same sense of the noun board: “Room and board .” and “He nailed boards across the windows .”? Return your answer as a letter: "T" if the sense is the same or "F" if it’s not the same. \\
        \vspace{-1mm}
        \textbf{\textsc{Model Answer \color{darkGreen}{(correct)}:}}expected: F, answer: F \\
        \midrule
        \vspace{-2mm}
        \textbf{\textsc{Question:}} Are the given texts expressing the same sense of the verb circulate: “Circulate a rumor .” and “This letter is being circulated among the faculty .”? Return your answer as a letter: "T" if the sense is the same or "F" if it’s not the same. \\
        \vspace{-1mm}
        \textbf{\textsc{Model Answer \color{blue}{(incorrect)}:}} expected: F, answer: T \\
        \midrule
        \vspace{-2mm}
        \textbf{\textsc{Question:}}Having the sentence "Always before, Larry had helped Dad with his work. But he could not help him now, for Dad said that his boss at the railroad company would not want anyone but him to work in the office." tell me if the following sentence "He could not help Larry now." is true or false? Answer a number "0" if false or "1" if true. \\
        \vspace{-1mm}
        \textbf{\textsc{Model Answer \color{darkGreen}{(correct but wrong format)}:}} expected: 0, answer: False. The sentence should actually say "Dad could not help Larry now." \\
        \midrule
        \vspace{-2mm}
        \textbf{\textsc{Question:}} Having the sentence "There is a pillar between me and the stage, and I can't see it." tell me if the following sentence "I can't see around the stage." is true or false? Answer a number "0" if false or "1" if true. \\
        \vspace{-1mm}
        \textbf{\textsc{Model Answer \color{blue}{(incorrect)}:}} expected: 1, answer: 0 \\
        \midrule
        \vspace{-2mm}
        \textbf{\textsc{Question:}}Having premise "Nokia, Texas Instruments and other leading makers of mobile phones have formally complained to Brussels that Qualcomm, the US mobile chipmaker, has unfairly used its patents on 3G technologies." judge if the following hypothesis "Texas Instruments produces mobile phones." are logically connected with the premise?           Answer "entailment" if yes, or "not\_entailment" if no. \\
        \vspace{-1mm}
        \textbf{\textsc{Model Answer \color{darkGreen}{(correct)}:}} expected: entailment, answer: entailment \\
        \bottomrule
    \end{tabular}
        \caption{
   WordContext, textEntail and WNLI task examples.
    }
    \label{tab:appendix-wordContext-WNLI-TextEntail-analysis}
\end{table*}
\endgroup

\begingroup
\begin{table*}[ht]

    \centering
    \small
    \begin{tabular}{p{\linewidth}}
        \toprule
        \vspace{-2mm}
        \textbf{\textsc{Question:}} The context is'In 2006, a toxic waste spill off the coast of Côte d'Ivoire, from a European ship, prompted the Commission to look into legislation against toxic waste. Environment Commissioner Stavros Dimas stated that "Such highly toxic waste should never have left the European Union". With countries such as Spain not even having a crime against shipping toxic waste, Franco Frattini, the Justice, Freedom and Security Commissioner, proposed with Dimas to create criminal sentences for "ecological crimes". The competence for the Union to do this was contested in 2005 at the Court of Justice resulting in a victory for the Commission. That ruling set a precedent that the Commission, on a supranational basis, may legislate in criminal law – something never done before. So far, the only other proposal has been the draft intellectual property rights directive. Motions were tabled in the European Parliament against that legislation on the basis that criminal law should not be an EU competence, but was rejected at vote. However, in October 2007, the Court of Justice ruled that the Commission could not propose what the criminal sanctions could be, only that there must be some.'. Give the result in JSON format with keys 'text' and 'answer\_start' with values: Answer to the question 'In which year did a toxic waste spill from a European ship prompt the Commission to look into legislation against waste?' in 1 words, the number of character at which the answer appears in the context. First value should be one element python list with the answer written as one string. Second value should be one element python list. \\
        \vspace{-1mm}
        \textbf{\textsc{Model Answer \color{darkGreen}{(correct)}:}} expected: \{'text': ['2006', '2006', '2006'], 'answer\_start': [3, 3, 3]\}, answer: \{     "text": ["2006"],     "answer\_start": [3] \} \\
        \midrule
        \vspace{-2mm}
        \textbf{\textsc{Question:}} The context is'Today, roller skating is easy and fun. But a long time ago, it wasn't easy at all. Before 1750, the idea of skating didn't exist. That changed because of a man named Joseph Merlin. Merlin's work was making musical instruments. In his spare time he liked to play the violin. Joseph Merlin was a man of ideas and dreams. People called him a dreamer.  One day Merlin received an invitation to attend a fancy dress ball. He was very pleased and a little excited. As the day of the party came near, Merlin began to think how to make a grand entrance at the party. He had an idea. He thought he would get a lot of attention if he could skate into the room.  Merlin tried different ways to make himself roll. Finally, he decided to put two wheels under each shoe. These were the first roller skates. Merlin was very proud of his invention and dreamed of arriving at the party on wheels while playing the violin.  On the night of the party Merlin rolled into the room playing his violin. Everyone was astonished to see him. There was just one problem. Merlin had no way to stop his roller skates. He rolled on and on. Suddenly, he ran into a huge mirror that was hanging on the wall. Down fell the mirror, breaking to pieces. Nobody forgot Merlin's grand entrance for a long time!'. Give correct answer to the question: 'People thought Merlin was a dreamer because he \_ .' from A, B, C, D where A, B, C, D correspond to each item in python list ['often gave others surprises', 'was a gifted musician', 'was full of imagination', 'invented the roller skates']. The answer must be only one letter. \\
        \vspace{-1mm}
        \textbf{\textsc{Model Answer \color{darkGreen}{(correct)}:}} expected: C, answer: C \\
        \midrule
        \vspace{-2mm}
        \textbf{\textsc{Question:}} The context is'Children have their own rules in playing games. They seldom need a judge and rarely trouble to keep scores. They don't care much about who wins or loses, and it doesn't seem to worry them if the game is not finished. Yet, they like games that depend a lot on luck, so that their personal abilities cannot be directly compared. They also enjoy games that move in stages, in which each stage, the choosing of leaders, the picking-up of sides, or the determining of which side shall start, is almost a game in itself.  Grown-ups can hardly find children's games exciting, and they often feel puzzled at why their kids play such simple games again and again. However, it is found that a child plays games for very important reasons. He can be a good player without having to think whether he is a popular person, and he can find himself being a useful partner to someone of whom he is ordinarily afraid. He becomes a leader when it comes to his turn. He can be confident, too, in particular games, that it is his place to give orders, to pretend to be dead, to throw a ball actually at someone, or to kiss someone he caught.  It appears to us that when children play a game they imagine a situation under their control. Everyone knows the rules, and more important, everyone plays according to the rules. Those rules may be childish, but they make sure that every child has a chance to win.'. Give correct answer to the question: 'To become a leader in a game the child has to \_ .' from A, B, C, D where A, B, C, D correspond to each item in python list ['play well', 'be popular among his playmates', 'be confident in himself', 'wait for his turn']. The answer must be only one letter. \\
        \vspace{-1mm}
        \textbf{\textsc{Model Answer \color{blue}{(incorrect and wrong format)}:}} expected: D, answer: C (be confident in himself) \\
        \bottomrule
    \end{tabular}
        \caption{
   SQuAD and ReAding task examples.
    }
    \label{tab:appendix-SQuAD-ReAding-analysis}
\end{table*}
\endgroup

\begingroup
\begin{table*}[ht]

    \centering
    \small
    \begin{tabular}{p{\linewidth}}
        \toprule
        \vspace{-2mm}
        \textbf{\textsc{Question:}} From the given list of all emotions, choose the ones that the input text arouses in most people reading it. Write your answer in the form of a Python list containing exactly 1 selected most matching emotion. List of all emotions: admiration, amusement, anger, annoyance, approval, caring, confusion, curiosity, desire, disappointment, disapproval, disgust, embarrassment, excitement, fear, gratitude, grief, joy, love, nervousness, optimism, pride, realization, relief, remorse, sadness, surprise, neutral. Input text: Well, there's cubs and otters too. \\
        \vspace{-1mm}
        \textbf{\textsc{Model Answer \color{darkGreen}{(correct)}:}}expected: neutral, answer: ["neutral"] \\
        \midrule
        \vspace{-2mm}
        \textbf{\textsc{Question:}} From the given list of all emotions, choose the ones that the input text arouses in most people reading it. Write your answer in the form of a Python list containing exactly 1 selected most matching emotion. List of all emotions: admiration, amusement, anger, annoyance, approval, caring, confusion, curiosity, desire, disappointment, disapproval, disgust, embarrassment, excitement, fear, gratitude, grief, joy, love, nervousness, optimism, pride, realization, relief, remorse, sadness, surprise, neutral. Input text: I would try rubbing alcohol, high percentage. Looks greasy to me. \\
        \vspace{-1mm}
        \textbf{\textsc{Model Answer \color{blue}{(incorrect)}:}} expected: optimism, answer: ["disgust"] \\
        \midrule
        \vspace{-2mm}
        \textbf{\textsc{Question: }}From the given list of all emotions, choose the ones that the input text arouses in most people reading it. List of all emotions: admiration, amusement, anger, annoyance, approval, caring, confusion, curiosity, desire, disappointment, disapproval, disgust, embarrassment, excitement, fear, gratitude, grief, joy, love, nervousness, optimism, pride, realization, relief, remorse, sadness, surprise, neutral. Text: UN report says population before war: 300,000 [NAME] population before war: 2,000,000. Write your answer in the form of a Python list containing exactly 1 unique selected most matching emotion. Do not explain yourself. \\
        \vspace{-1mm}
        \textbf{\textsc{Model Answer \color{blue}{(incorrect and out of range)}:}} expected: neutral, answer: ["disbelief"] \\
        \midrule
        \vspace{-2mm}
        \textbf{\textsc{Question:}} Knowing that a certain person described known texts with provided emotions: Known text: Honestly if I didn't have this many favorite characters not yet in the game, I would probably waste my 7 votes on Merlinus.; Emotion: love. Known text: The Japanese and the British can do it well, but they probably get like 10x the funding per capita.; Emotion: pride With what emotions would this person describe the unknown text? List of all emotions: admiration, amusement, anger, annoyance, approval, caring, confusion, curiosity, desire, disappointment, disapproval, disgust, embarrassment, excitement, fear, gratitude, grief, joy, love, nervousness, optimism, pride, realization, relief, remorse, sadness, surprise, neutral. Unknown text: the most romantic thing that can possibly happen to me. Write your answer in the form of a Python list containing exactly 1 unique selected most matching emotion. Do not explain yourself. \\
        \vspace{-1mm}
        \textbf{\textsc{Model Answer \color{darkGreen}{(correct)}:}} expected: love, answer: ["love"] \\
        \bottomrule
    \end{tabular}
        \caption{
   GoEmo and GoEmoPer task examples.
    }
    \label{tab:appendix-GoEmo-analysis}
\end{table*}
\endgroup

\begingroup
\begin{table*}[ht]

    \centering
    \small
    \begin{tabular}{p{\linewidth}}
        \toprule
        \vspace{-2mm}
        \textbf{\textsc{Question:}} Describe the sentiment of the given text. Choose your answer from provided list and
map your answer with following negative: 0, neutral: 1, positive: 2 and return an integer as a result. Do not explain yourself.
Input data:
Text: An interesting security vulnerability - albeit not for the everyday car thief
Possible sentiment: negative, neutral, positive
 \\
        \vspace{-1mm}
        \textbf{\textsc{Model Answer \color{darkGreen}{(correct)}:}} expected: 1, answer: 1 \\
        \midrule
        \vspace{-2mm}
        \textbf{\textsc{Question:}} Describe the sentiment of the given text. Choose your answer from provided list and
map your answer with following negative: 0, neutral: 1, positive: 2 and return an integer as a result. Do not explain yourself.
Input data:
Text: Can someone tell me where I can get a good red eyeshadow from?? (Not animal testing brands pls x)
Possible sentiment: negative, neutral, positive
\\
        \vspace{-1mm}
        \textbf{\textsc{Model Answer \color{darkGreen}{(correct but wrong format)}:}} expected: 1, answer: Neutral: 1 \\
        \midrule
        \vspace{-2mm}
        \textbf{\textsc{Question:}} 
Assuming that you have to describe the stance from the given list which is evoked after 
reading the text about Hillary Clinton by majority of people, which one would you select? 
Map your answer with following none:  0, against: 1, favor: 2. Do not explain yourself.
Input data:

Text: President Jimmy Carter:  There's no doubt Hillary will get the nomination. And when she does I'll be happy to support her \#SemST

Possible stances: none, against, favor

 \\
        \vspace{-1mm}
        \textbf{\textsc{Model Answer \color{darkGreen}{(correct)}:}} expected: 2, answer: 2 \\
        \bottomrule
    \end{tabular}
        \caption{
   TweetEmoji, TweetSent and TweetStance task examples.
    }
    \label{tab:appendix-TweetEval-analysis}
\end{table*}
\endgroup

\clearpage

\section{Full Feature Analysis}
\label{appendix:full_feature_analyze}
In this section, we first introduce the feature details of our feature extraction process to complete the full category in Table~\ref{tab:used_tools}. Then we will show full knowledge and linguistic feature analysis results of representative feature for each category. In section~\ref{sec:temporal_analysis}, we just show 3 features (WRich\_S)  in AdSem category, actually there are 9 categories of total 265 features.

\subsection{Feature Meaning}
\label{appendix:feature_meaning}
We list representative features of each category in the following Tables. Table ~\ref{Append1} to \ref{Append5} introduce the linguistic feature categories. As those linguistic features are extracted by the LingFeat\footnote{\url{https://github.com/brucewlee/lingfeat}} tool, ~\citet{lee-etal-2021-pushing} has introduced the entire 255 linguistic features in their paper, please refer to their paper's appendix for full linguistic feature information. On the other hand, we also extract 10 knowledge features using UIE~\cite{lu-etal-2022-unified} and CogIE~\cite{jin-etal-2021-cogie} tools.  Features in Table ~\ref{Append6} and \ref{Append7} are named entities and opinions extracted by UIE\footnote{\url{https://github.com/universal-ie/UIE}}. And features in Table ~\ref{Append8} and Table \ref{Append9} are entity relations and semantic frames extracted by CogIE\footnote{\url{https://github.com/jinzhuoran/CogIE}}.  

\begin{table}[H]
    \centering
    \resizebox{0.5\textwidth}{!}{%
    \begin{tabular}{l@{\hspace{0.3ex}}|l@{\hspace{0.5ex}}|l@{\hspace{0.5ex}}}
    \hline
        \textbf{$idx$} & \textbf{Code} & \textbf{Definition} \\ \hline
        1              & WRich05\_S     &Richness, 50 topics extracted from Wikipedia Dump        \\
        2              & WClar05\_S     &Clarity, 50 topics extracted from Wikipedia Dump         \\
        3              & WNois05\_S     &Noise, 50 topics extracted from Wikipedia Dump           \\
        4              & WTopc05\_S     &\# of topics, 50 topics extracted from Wikipedia Dump        \\
               5              & WRich10\_S     &Richness, 100 topics extracted from Wikipedia Dump      \\
        6              & WClar10\_S     &Clarity, 100 topics extracted from Wikipedia Dump       \\
        7              & WNois10\_S     &Noise, 100 topics extracted from Wikipedia Dump         \\
        8              & WTopc10\_S     &\# of topics, 100 topics extracted from Wikipedia Dump       \\
        9              & WRich15\_S     &Richness, 150 topics extracted from Wikipedia Dump      \\
        10             & WClar15\_S     &Clarity, 150 topics extracted from Wikipedia Dump       \\
                ... & ... & ... \\
        20             & BTopc05\_S     &\# of topics, 50 topics extracted from WeeBit Corpus       \\         
          21             & BRich10\_S     &Richness, 100 topics extracted from WeeBit Corpus     \\         
        22             & BClar10\_S     &Clarity, 100 topics extracted from WeeBit Corpus      \\        
        23             & BNois10\_S     &Noise, 100 topics extracted from WeeBit Corpus        \\        
        24             & BTopc10\_S     &\# of topics, 100 topics extracted from WeeBit Corpus      \\        
        25             & BRich15\_S     &Richness, 150 topics extracted from WeeBit Corpus     \\         
        26             & BClar15\_S     &Clarity, 150 topics extracted from WeeBit Corpus      \\        
        27             & BNois15\_S     &Noise, 150 topics extracted from WeeBit Corpus        \\        
        28             & BTopc15\_S     &\# of topics, 150 topics extracted from WeeBit Corpus      \\        
        29             & BRich20\_S     &Richness, 200 topics extracted from WeeBit Corpus     \\         
        30             & BClar20\_S     &Clarity, 200 topics extracted from WeeBit Corpus      \\         
        32             & BTopc20\_S     &\# of topics, 200 topics extracted from WeeBit Corpus      \\ \hline         
        33             & ORich05\_S     &Richness, 50 topics extracted from OneStop Corpus    \\          
        34             & OClar05\_S     &Clarity, 50 topics extracted from OneStop Corpus     \\           
        35             & ONois05\_S     &Noise, 50 topics extracted from OneStop Corpus       \\           
        36             & OTopc05\_S     &\# of topics, 50 topics extracted from OneStop Corpus     \\           
        ... & ... & ... \\       
        48             & OTopc20\_S     &\# of topics, 200 topics extracted from OneStop Corpus    \\ \hline                
    \end{tabular}
    }
\caption{\label{Append1}Advanced Semantic(AdSem) feature category, including Wikipedia Knowledge Features (W), WeeBit Knowledge Features (B), OneStop Knowledge Features (O).}
\end{table}

\begin{table}[H]
    \centering
    \resizebox{0.5\textwidth}{!}{%
    \begin{tabular}{l@{\hspace{0.3ex}}|l@{\hspace{0.5ex}}|l@{\hspace{0.5ex}}}
    \hline
        \textbf{$idx$} & \textbf{Code} & \textbf{Definition} \\ \hline
        49             & to\_EntiM\_C    & total number of Entities Mentions                       \\              
        50             & as\_EntiM\_C    & average number of Entities Mentions per sentence        \\              
        51             & at\_EntiM\_C    & average number of Entities Mentions per token (word)           \\              
        52             & to\_UEnti\_C    & total number of unique Entities                                \\              
        53             & as\_UEnti\_C    & average number of unique Entities per sentence                 \\              
        54             & at\_UEnti\_C    & average number of unique Entities per token (word)             \\ \hline         
  55             & ra\_SSToT\_C   & ratio of SS transitions $:$ total, count from Entity Grid         \\
        56             & ra\_SOToT\_C   & ratio of SO transitions $:$ total, count from Entity Grid         \\
        57             & ra\_SXToT\_C   & ratio of SX transitions $:$ total, count from Entity Grid         \\
        58             & ra\_SNToT\_C   & ratio of SN transitions $:$ total, count from Entity Grid         \\
        
       ... & ... & ... \\  
        72             & LoCohPW\_S     & Local Coherence for PW score from Entity Grid          \\
        73             & LoCohPU\_S     & Local Coherence for PU score from Entity Grid          \\
        74             & LoCoDPA\_S     & Local Coherence dist. for PA score from Entity Grid \\
        75             & LoCoDPW\_S     & Local Coherence dist. for PW score from Entity Grid \\
        76             & LoCoDPU\_S     & Local Coherence dist. for PU score from Entity Grid  \\ \hline
    \end{tabular}
    }
\caption{\label{Append2}Discourse (Disco) feature category, including Entity Density Features (EnDF). Entity Grid Features (EnGF).}
\end{table}

\begin{table}[H]
    \centering
    \resizebox{0.5\textwidth}{!}{%
    \begin{tabular}{l@{\hspace{0.5ex}}|l@{\hspace{0.5ex}}|l@{\hspace{0.8ex}}}
    \hline
        \textbf{$idx$} & \textbf{Code}  & \textbf{Definition} \\ \hline
        77     & to\_NoPhr\_C   & total count of Noun phrases                                     \\
        78     & as\_NoPhr\_C   & average count of Noun phrases per sentence                      \\
        79     & at\_NoPhr\_C   & average count of Noun phrases per token                         \\
        80     & ra\_NoVeP\_C   & ratio of Noun phrases : Verb phrases count               \\
        ... & ... & ... \\  
        124    & ra\_AvAjP\_C   & ratio of Adv phrases : Adj phrases count                 \\    \hline      
        
    125   & to\_TreeH\_C   & total parsed Tree Height of all sentences                              \\
    126   & as\_TreeH\_C   & average parsed Tree Height per sentence                                \\
    127   & at\_TreeH\_C   & average parsed Tree Height per token                           \\
    128   & to\_FTree\_C   & total length of Flattened parsed Trees                                 \\
    129   & as\_FTree\_C   & average length of Flattened parsed Trees per sentence                  \\
    130   & at\_FTree\_C   & average length of Flattened parsed Trees per token              \\     \hline   
 131   & to\_NoTag\_C   & total count of Noun tags                                                   \\
    132   & as\_NoTag\_C   & average count of Noun tags per sentence                                    \\
    133   & at\_NoTag\_C   & average count of Noun tags per token                                       \\
   
     ... & ... & ... \\  
    185   & ra\_CoFuW\_C   & ratio of Content words to Function words               \\ \hline             
    
    \end{tabular}
    }
\caption{\label{Append3}Syntactic (Synta) feature category, including Phrasal Features (PhrF), Tree Structural Features (TrSF), Part-of-Speech Features (POSF)}
\end{table}

\begin{table}[H]
    \centering
    \resizebox{0.5\textwidth}{!}{%
    \begin{tabular}{l@{\hspace{0.3ex}}|l@{\hspace{0.3ex}}|l@{\hspace{0.3ex}}}
    \hline
        \textbf{$idx$} & \textbf{Code}  & \textbf{Definition} \\ \hline
    186   & SimpNoV\_S    & unique Nouns/total Nouns \#Noun Variation                   \\
    187   & SquaNoV\_S    & (unique Nouns**2)/total Nouns \#Squared Noun Variation      \\
    188   & CorrNoV\_S    & unique Nouns/sqrt(2*total Nouns) \#Corrected Noun Var \\
    ... & ... & ... \\  
    197   & CorrAvV\_S    & unique Adv/sqrt(2*total Adv) \#Corrected Adv Var      \\ \hline  
     198   & SimpTTR\_S    & unique tokens/total tokens \#TTR                                     \\
    199   & CorrTTR\_S    & unique/sqrt(2*total) \#Corrected TTR                   \\
    200   & BiLoTTR\_S    & log(unique)/log(total) \#Bi-Logarithmic TTR            \\
    201   & UberTTR\_S    & (log(unique))$^2$/log(total/unique) \#Uber  \\
    202   & MTLDTTR\_S    & \#Measure of Textual Lexical Diversity (TTR, 0.72)            \\   \hline
    203   & to\_AAKuW\_C   & total AoA (Age of Acquisition) of words, Kuperman                       \\
    204   & as\_AAKuW\_C   & average AoA of words per sentence, Kuperman                              \\
    205   & at\_AAKuW\_C   & average AoA of words per token, Kuperman                                 \\
    ... & ... & ... \\  
    217   & at\_AACoL\_C   & average AoA of lem, Cortese and K norm per token       \\ \hline
    218   & to\_SbFrQ\_C   & total SubtlexUS FREQcount value                  \\
    219   & as\_SbFrQ\_C   & average SubtlexUS FREQcount value per sentence    \\
    220   & at\_SbFrQ\_C   & average SubtlexUS FREQcount value per token      \\
    ... & ... & ... \\ 
    241   & at\_SbL1C\_C   & average SubtlexUS Lg10CD value per token         \\ \hline
    \end{tabular}
    }
\caption{\label{Append4} Lexico Semantic (LxSem) feature category, including Variation Ratio Features (VarF), Type Token Ratio Features (TTRF), Psychollinguistic Features (PsyF) and Word Familiarity Features (WorF).}
\end{table}

\begin{table}[H]
    \centering
    \resizebox{0.5\textwidth}{!}{%
    \begin{tabular}{l@{\hspace{0.3ex}}|l@{\hspace{0.3ex}}|l@{\hspace{0.3ex}}}
    \hline
        \textbf{$idx$} & \textbf{Code}  & \textbf{Definition} \\ \hline      
    242   & TokSenM\_S   & total count of tokens x total count of sentence          \\                      
    243   & TokSenS\_S   & sqrt(total count of tokens x total count of sentence)    \\                         
    244   & TokSenL\_S   & log(total count of tokens)/log(total count of sent)  \\                       
    245   & as\_Token\_C  & average count of tokens per sentence                     \\                      
    246   & as\_Sylla\_C  & average count of syllables per sentence                  \\                      
    247   & at\_Sylla\_C  & average count of syllables per token                     \\                      
    248   & as\_Chara\_C  & average count of characters per sentence                 \\                      
    249   & at\_Chara\_C  & average count of characters per token                    \\ \hline    
    250   & SmogInd\_S    & Smog Index                        \\
251   & ColeLia\_S    & Coleman Liau Readability Score   \\
252   & Gunning\_S    & Gunning Fog Count Score (New, US Navy Report)           \\
253   & AutoRea\_S    & Automated Readability Idx (New, US Navy Report)   \\ 
254   & FleschG\_S    & Flesch Kincaid Grade Level (New, US Navy Report)        \\ 
255   & LinseaW\_S    & Linsear Write Formula Score       \\ \hline
    \end{tabular}
    }
\caption{\label{Append5} Shallow Feature (ShaF) category.}
\end{table}

\begin{table}[H]
    \centering
    \resizebox{0.5\textwidth}{!}{%
    \begin{tabular}{l@{\hspace{0.3ex}}|l@{\hspace{0.3ex}}|l@{\hspace{0.3ex}}}
    \hline
        \textbf{$idx$} & \textbf{Code}  & \textbf{Definition} \\ \hline      
    256   & Person  & total count of Person entities          \\                      
    257   & Organization   & total count of Organization entities     \\                         
    258   & Location  & total count of Location entities  \\                       
    259   & Work & total count of Work entities                    \\                      
    260   & Time & total count of Time entities                 \\                                       
 \hline
    \end{tabular}
    }
\caption{\label{Append6} Named entity (NE) feature category.}
\end{table}

\begin{table}[H]
    \centering
    \resizebox{0.5\textwidth}{!}{%
    \begin{tabular}{l@{\hspace{0.3ex}}|l@{\hspace{0.3ex}}|l@{\hspace{0.3ex}}}
    \hline
        \textbf{$idx$} & \textbf{Code}  & \textbf{Definition} \\ \hline      
    261   & Opinion & total count of Opinion expression words                 \\
        262   & Sentiment &  A probability for negative, neutral or positive sentiment \\
 \hline
    \end{tabular}
    }
\caption{\label{Append7} Opinion (OP) feature category.}
\end{table}

\begin{table}[H]
    \centering
    \resizebox{0.5\textwidth}{!}{%
    \begin{tabular}{l@{\hspace{0.3ex}}|l@{\hspace{0.3ex}}|l@{\hspace{0.3ex}}}
    \hline
        \textbf{$idx$} & \textbf{Code}  & \textbf{Definition} \\ \hline      
    263   & Relation & total count of entity relations                \\
 \hline
    \end{tabular}
    }
\caption{\label{Append8} Relation Extraction (RE) feature category. Actually there are 500 kinds of relation classes in CogIE tool that we use.}
\end{table}

\begin{table}[H]
    \centering
    \resizebox{0.5\textwidth}{!}{%
    \begin{tabular}{l@{\hspace{0.3ex}}|l@{\hspace{0.3ex}}|l@{\hspace{0.3ex}}}
    \hline
        \textbf{$idx$} & \textbf{Code}  & \textbf{Definition} \\ \hline      
    264   & Lexical Unit & total count of evoking words              \\
        265   & Frame Element &  total count of semantic role words  \\
 \hline
    \end{tabular}
    }
\caption{\label{Append9}Frame-Semantic Parsing (FP) feature category. Actually there are 800 kinds of Lexical Units and 1900 Frame Elements in CogIE tool that we use. Lexical Unit is a word that plays the role of evoking the corresponding frame. Frame Element indicates different semantic roles associated with the frame.}
\end{table}

\newpage
\subsection{Knowledge Feature Analysis}
\label{appendix:knowledge_feature_analysis}

In this section, we will demonstrate the knowledge features on our ChatLog-Daily dataset to observe ChatGPT's patterns on knowledge. As knowledge features are mostly extracted mentions in sentences that can be linked with entity, relation or frame in the pre-defined knowledge schemas such as Wikipedia and FrameNet~\cite{baker-etal-1998-berkeley-framenet}, these knowledge are commonly string type rather than numerical type. Therefore, it is hard to comprehensively describe these string features with our defined numerical calculation. In this work, we limit the research scope by just using the count of these string type features to summarize, for example, using the count of relations in one document to summarize the relation category features. We leave this to future works for the exploration of more elaborate designs on knowledge features.

\begin{figure*}[htbp]
    \centering
    \includegraphics[width=\linewidth]{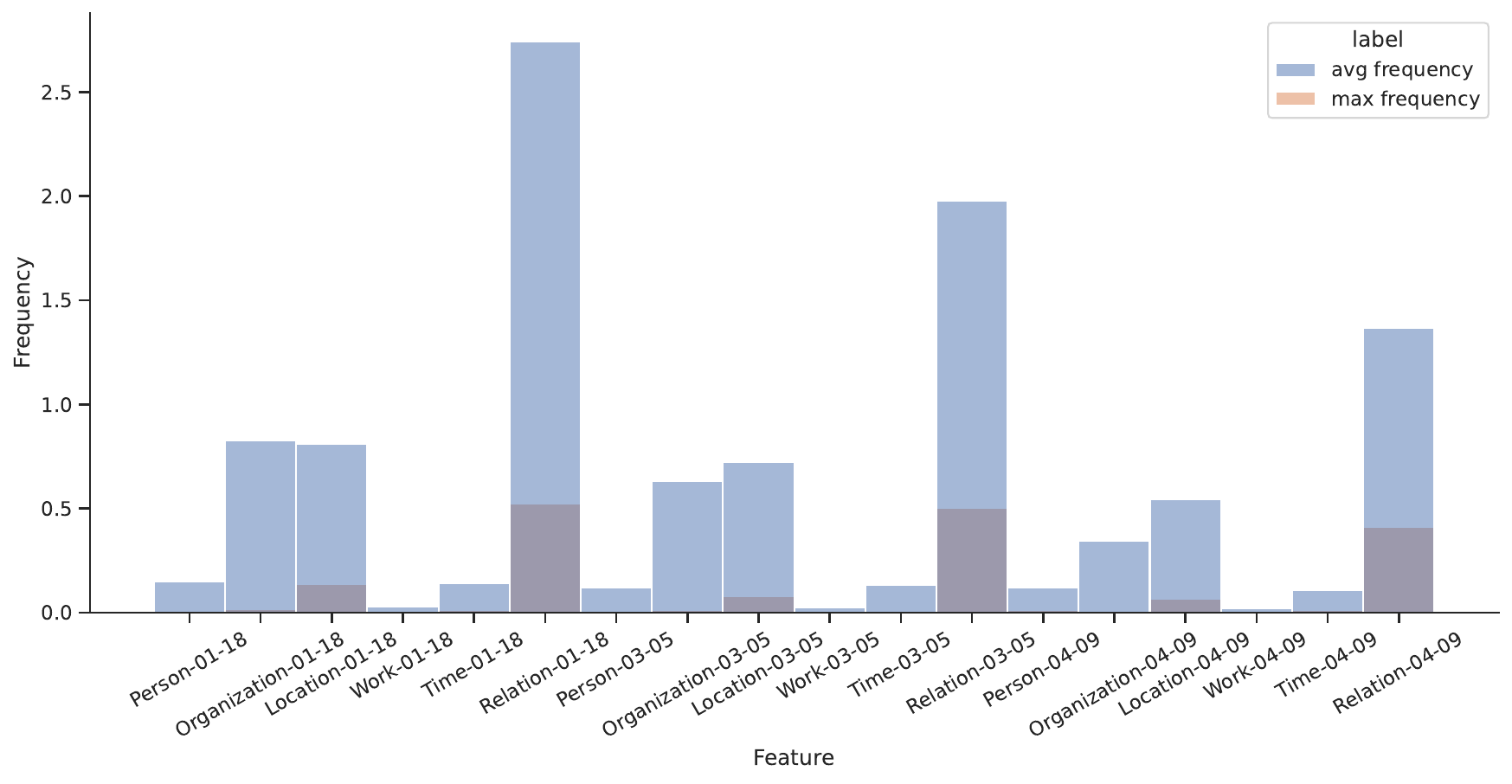}
    \caption{
    The extracted named entity and relation features' frequencies across time, where avg frequency denotes the average number of the feature in one ChatGPT answer and max frequency denotes the most common instance's frequency. Note that we demonstrate three dates' results in this figure, i.e. 2023-01-18, 2023-03-05 and 2023-04-09, which come from HC3 and ChatLog-Daily dataset respectively.
    }
    \label{fig:entity_and_relation}
\end{figure*}

\paragraph{Entity and Relation Analysis}
From the results presented in Figure ~\ref{fig:entity_and_relation}, we can observe some interesting patterns in ChatGPT's entity and relation features over time. Firstly, we note that there is an overall decrease in the frequency of named entity features in ChatGPT's answers, with the average frequency value for Organization feature decreasing from around 0.8 in January to around 0.4 in April. Additionally, the most common instance's frequency also shows an downward trend, suggesting that ChatGPT's ability to incorporate concise named entities in its responses is improving over time.

Similarly, we can observe an decrease in the frequency of the relation feature, with the average and maximum frequency value also showing an downward trend across the three dates presented. This result suggests that ChatGPT has become more effective and efficient in generating responses that contain relations between entities over time.

\paragraph{Opinion Analysis}

\begin{figure}[htbp]
    \centering
    \includegraphics[width=\linewidth]{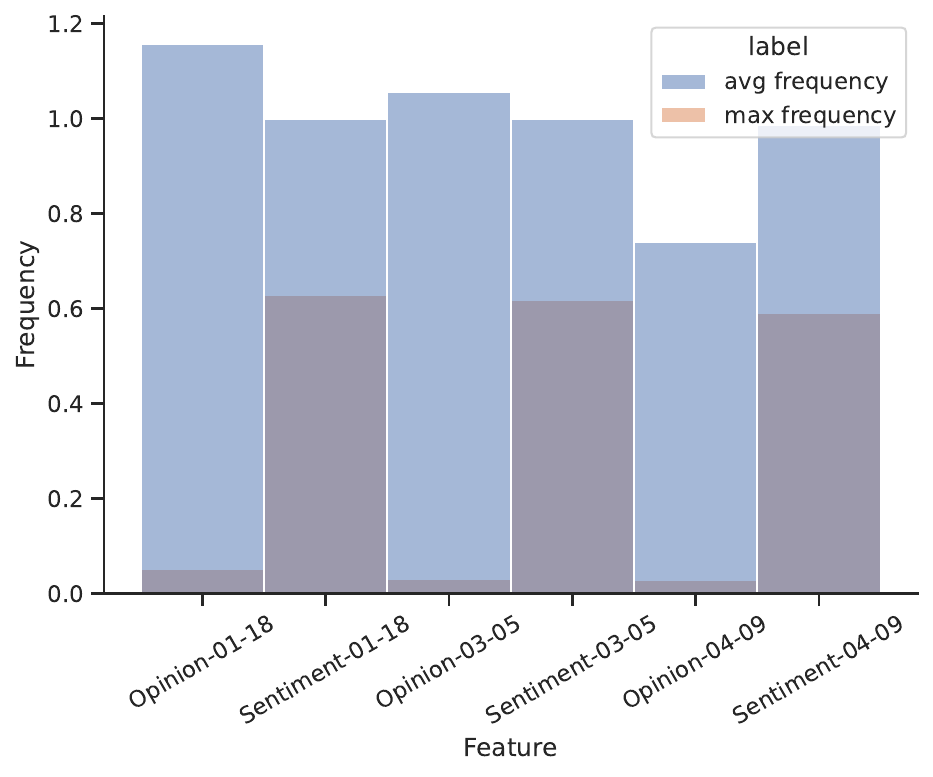}
    \caption{
    The extracted opinion and sentiment features' frequencies across time, where avg frequency denotes the average number of the feature in one ChatGPT answer and max frequency denotes the most common instance's frequency. Note that we demonstrate three dates' results in this figure, i.e. 2023-01-18, 2023-03-05 and 2023-04-09, which come from HC3 and ChatLog-Daily dataset respectively.
    }
    \label{fig:op_st}
\end{figure}
Figure ~\ref{fig:op_st} showcases the frequencies of opinion words and sentiment features extracted by ChatGPT. These features play a crucial role in understanding the intention that can make a significant difference in the text's overall meaning.

From the results presented in Figure ~\ref{fig:op_st}, we can see that the frequency of opinion words in ChatGPT's responses has decreased gradually over time. Specifically, the average frequency values for these features have decreased from 1.2 to 0.8. This suggests that ChatGPT has become more objective in generating  responses that express less opinions.

However, we can also see a unchanged value in the frequency of sentiment classification feature that indicating positive, negative, and neutral sentiment are stable in ChatGPT's responses over time. As shown in Table ~\ref{tab:topic_detail}, the most frequent sentiment is negative, maybe because that the questions are collected from Reddit, which are controversial and subjective. Actually there are almost the 40\% results are positive and neutral, which demonstrates a function of ChatGPT to generate responses that convey the correct sentiment.

\begin{figure}[htbp]
    \centering
    \includegraphics[width=\linewidth]{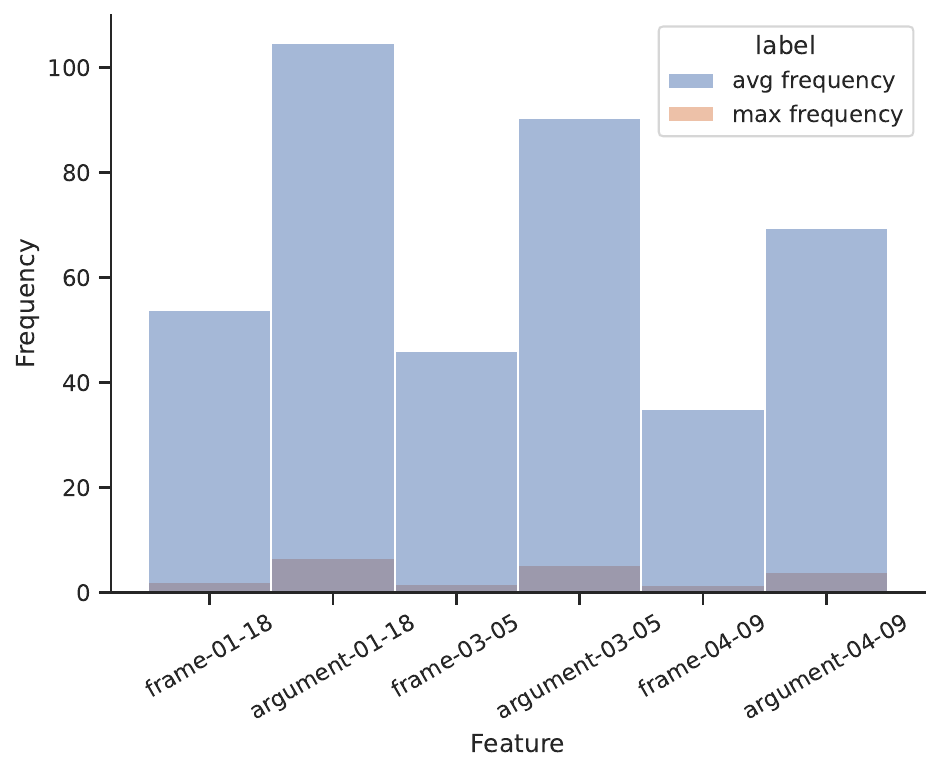}
    \caption{
    The extracted frame and argument features' frequencies across time, where avg frequency denotes the average number of the feature in one ChatGPT answer and max frequency denotes the most common instance's frequency. Note that we demonstrate three dates' results in this figure, i.e. 2023-01-18, 2023-03-05 and 2023-04-09, which come from HC3 and ChatLog-Daily dataset respectively.
    }
    \label{fig:fn_argument}
\end{figure}

\paragraph{Frame Analysis} From the results presented in Figure \ref{fig:fn_argument}, we observe an overall downward trend in the frequency of semantic frame words over time. This suggests that ChatGPT is becoming better at generating responses that contain semantic frames or frames used to represent entities and events. Similarly, we can observe an decreasing trend in the frequency of argument role features such as Agent, Patient, and Theme in ChatGPT's responses over time. This indicates an improvement in ChatGPT's ability to recognize the argument roles of entities and events in generating responses that convey the meaning correctly.

Overall, the results presented in Figure ~\ref{fig:entity_and_relation}, ~\ref{fig:op_st} and ~\ref{fig:fn_argument} demonstrate that ChatGPT's patterns on knowledge are showing improvement over time. The decreasing frequency values of named entity, relation, opinion and frame features suggest that ChatGPT has become more proficient in generating responses that contain less complex knowledge to fit the instruction  "Explain like I am five". Nonetheless, this study is limited in its scope of long-form QA task on ELI5~\cite{fan-etal-2019-eli5} dataset, and further research is required to explore more elaborate knowledge analysis for more tasks and to evaluate the generalizability of our findings.

\begin{table*}[htbp]
    \centering
    \begin{tabular}{|c|c|c c c|}
        \hline
        \multirow{2}{*}{\textbf{Type}} & \multirow{2}{*}{\textbf{Feature}} &\multicolumn{3}{c|}{\textbf{ Most Common Instances of 3 snapshots} }\\
        & & \textbf{ 2023-01-18} &  \textbf{ 2023-03-05} &  \textbf{ 2023-04-09}\\
        \hline 
        \multirow{6}{*}{\makecell*[c]{Entity\\ and \\ Relation}} & Person &  John Smith& Jesus& Jesus \\ \cline{2-5}
                                 & Organization & ISIS & Apple & Apple\\ \cline{2-5}
                                 & Location & United States& Earth& Earth\\ \cline{2-5}
                                 & Work & The Mona Lisa& work&  NY Times\\ \cline{2-5}
                                & Time & night& night& school hours\\ \cline{2-5}
                        & Relation & subclass of& shares border with& shares border with\\ \cline{2-5}
        \hline
        \multirow{2}{*}{Opinion}    & Opinion &  hope& want& want\\ \cline{2-5}
                                & Sentiment & negative& negative& negative \\ \cline{2-5}
        \hline
        \multirow{2}{*}{Frame}  & Frame & Causation& Causation& Causation \\ \cline{2-5}
                                 & Argument & entity& entity& entity \\ \cline{2-5}
        \hline
    \end{tabular}
    \caption{Detailed instance allocation of each knowledge feature. Note that we demonstrate three snapshots' most common instances respectively in this table, i.e. 2023-01-18, 2023-03-05 and 2023-04-09, which come from HC3 and ChatLog-Daily dataset.}
    \label{tab:topic_detail}
\end{table*}

\clearpage

\subsection{Linguistic Feature Analysis}
\label{appendix:linguistic_feature_analysis}
\begin{figure}[htbp]
    \centering
    \includegraphics[width=\linewidth]{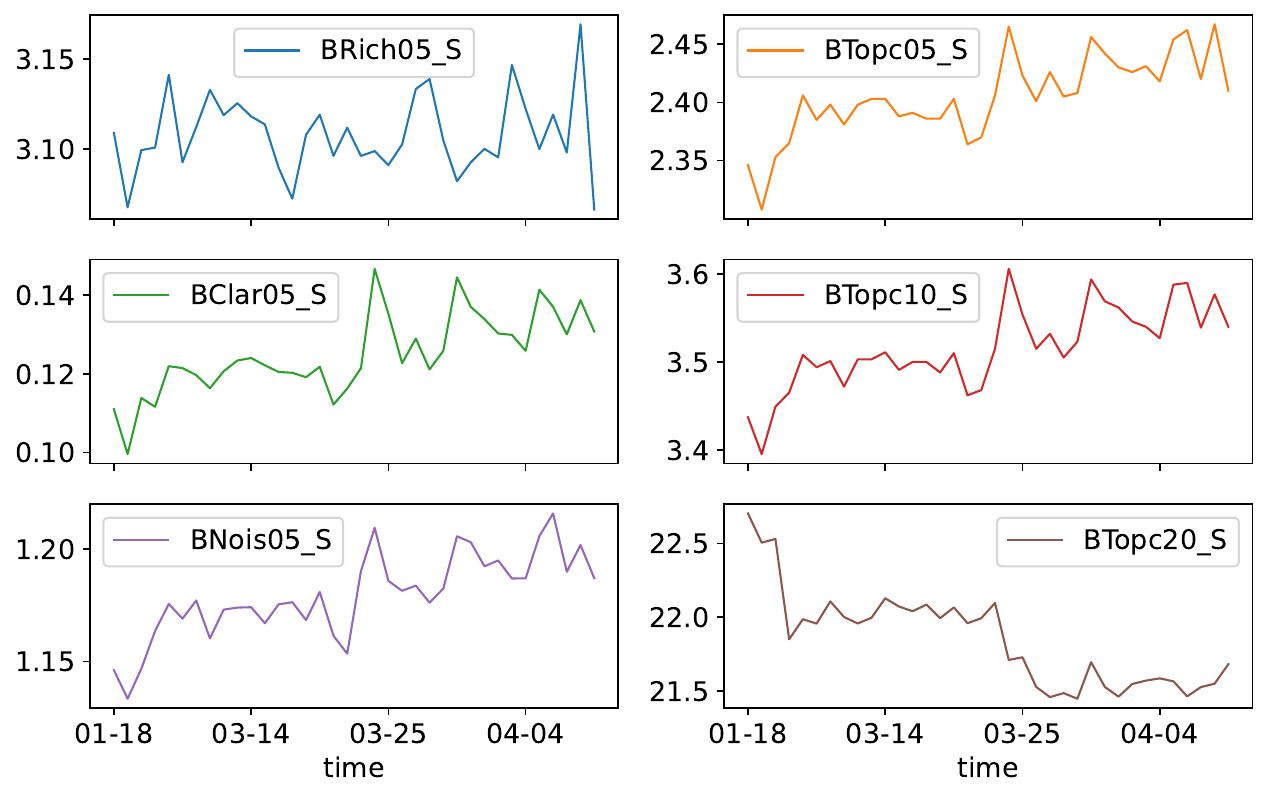}
    \caption{ 
   Trend of ChatGPT's Advanced Semantic (AdSem) linguistic features including Richness (BRich), Clarity (BClar), Noise (BNois) and number of topics(BTopc) scores on ChatLog-Daily dataset every day from March 5 to April 9, 2023. Note that the data at start point of January 18 comes from HC3 dataset. 
    }
    \label{fig:Adsem_v1}
\end{figure}

\begin{figure}[htbp]
    \centering
    \includegraphics[width=\linewidth]{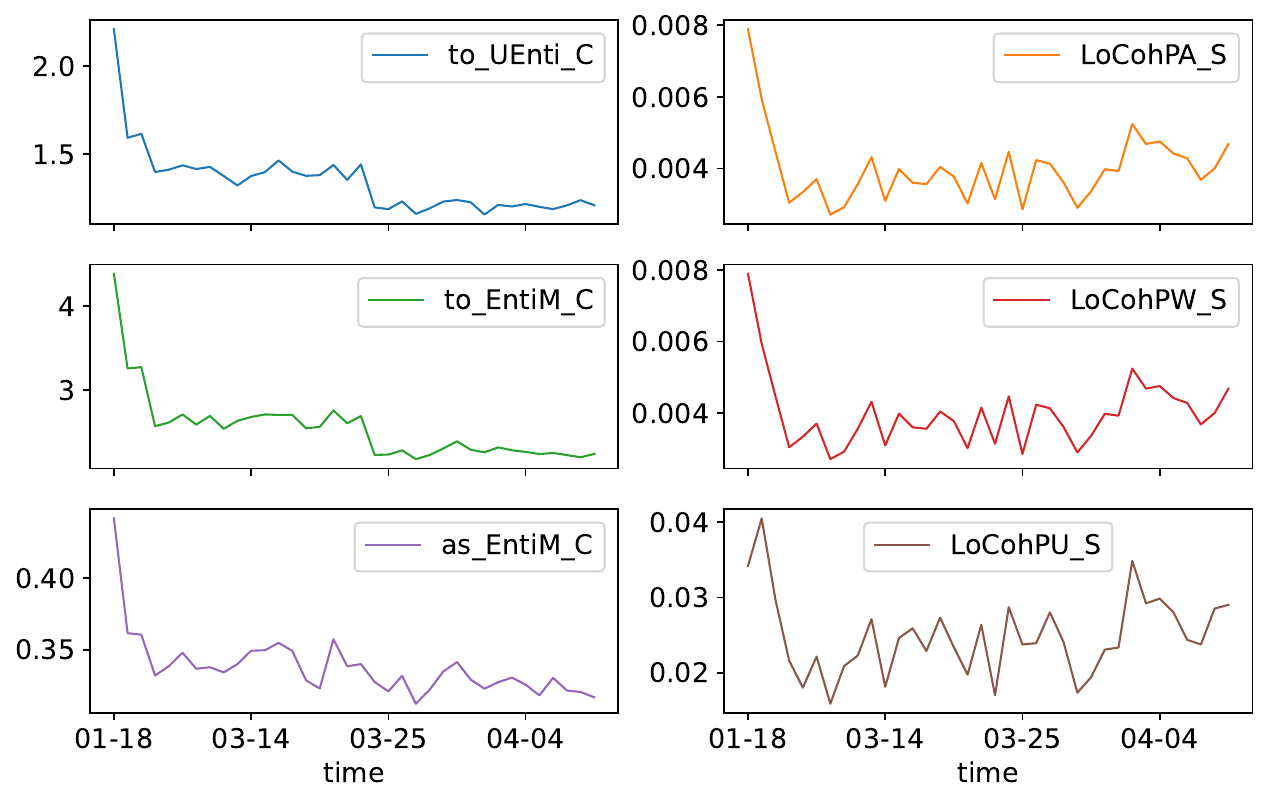}
    \caption{ 
   Trend of ChatGPT's Discourse (Disco) linguistic features including number of entities (Enti\_C) and Local Coherence from Entity Grid (LoCoh\_S) scores on ChatLog-Daily dataset every day from March 5 to April 9, 2023. Note that the data at start point of January 18 comes from open-access HC3 dataset. 
    }
    \label{fig:Disco_v1}
\end{figure}

\begin{figure}[htbp]
    \centering
    \includegraphics[width=\linewidth]{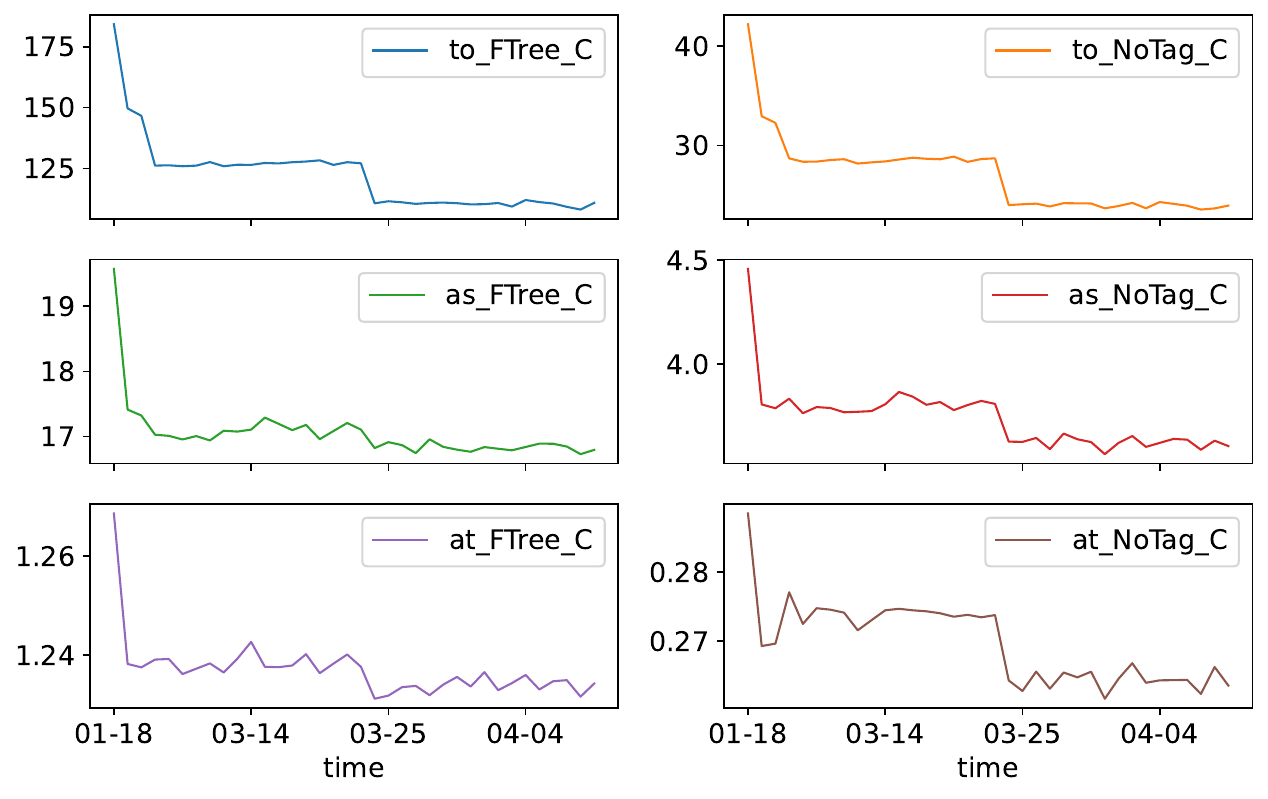}
    \caption{ 
   Trend of ChatGPT's Syntactic (Synta) linguistic features including length of Flattened parsed Trees (FTree\_C) and Local Coherence from count of Noun Tags (NoTag\_C) scores on ChatLog-Daily dataset every day from March 5 to April 9, 2023. Note that the data at start point of January 18 comes from open-access HC3 dataset. 
    }
    \label{fig:Synta_v1}
\end{figure}

\begin{figure}[htbp]
    \centering
    \includegraphics[width=\linewidth]{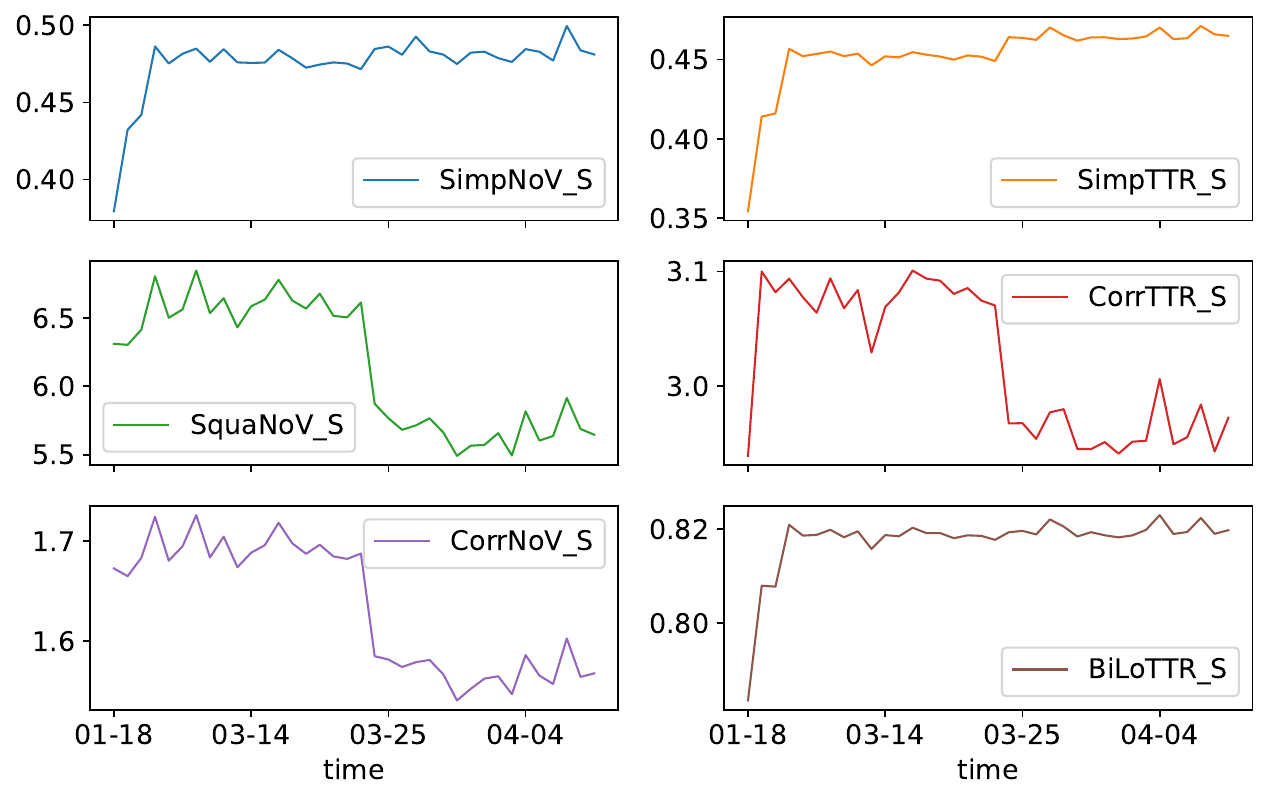}
    \caption{ 
     Trend of ChatGPT's Lexico Semantic (LxSem) linguistic features including the number of Noun Variation (NOV\_S) and the number of TTR (TTR\_S) scores on ChatLog-Daily dataset every day from March 5 to April 9, 2023. Note that the data at start point of January 18 comes from HC3 dataset. 
    }
    \label{fig:LxSem_v1}
\end{figure}

\begin{figure}[htbp]
    \centering
    \includegraphics[width=\linewidth]{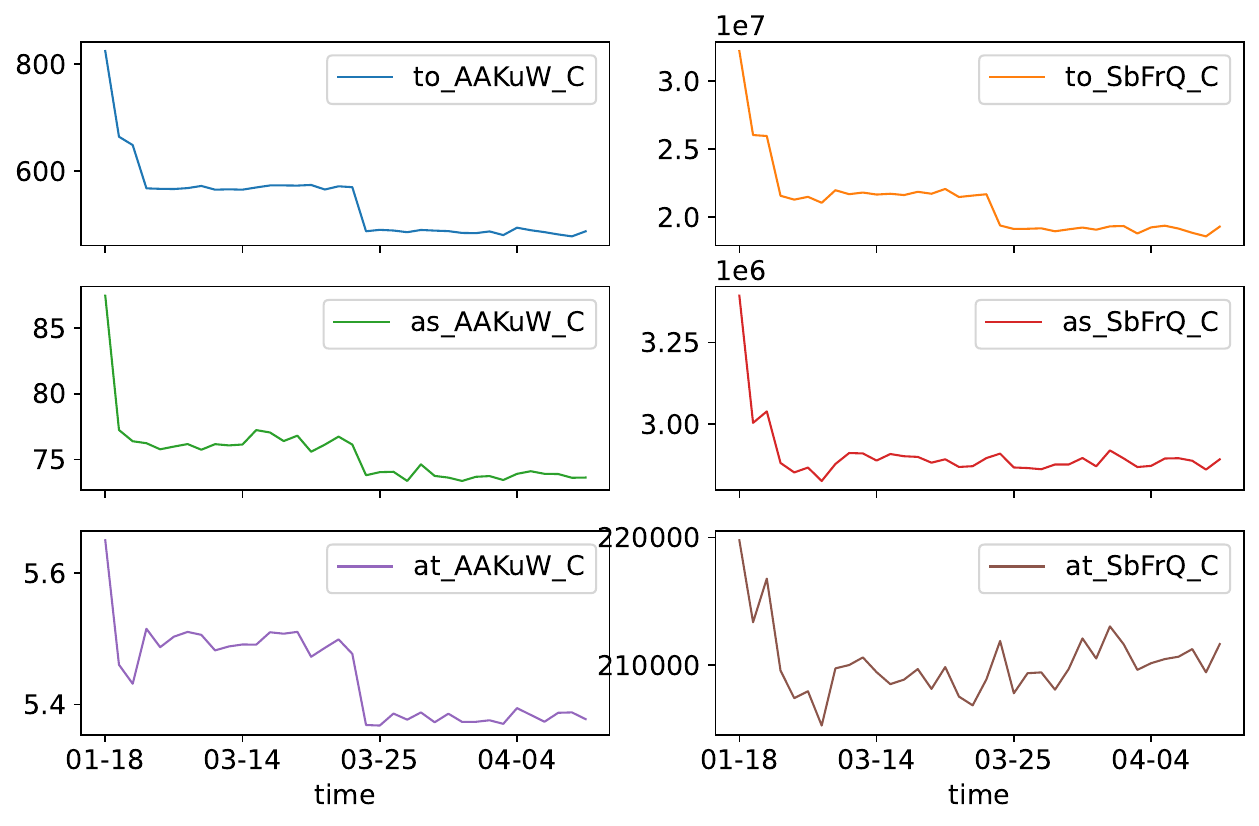}
    \caption{ 
   Trend of ChatGPT's Lexico Semantic (LxSem) linguistic features including the number of Age of Acquisition of words by Kuperman (AAKuW\_C) and the SubtleexUS FREQcount value (SbFrQ\_C) scores on ChatLog-Daily dataset every day from March 5 to April 9, 2023. Note that the data at start point of January 18 comes from HC3 dataset. 
    }
    \label{fig:LxSem_v2}
\end{figure}

\begin{figure}[htbp]
    \centering
    \includegraphics[width=\linewidth]{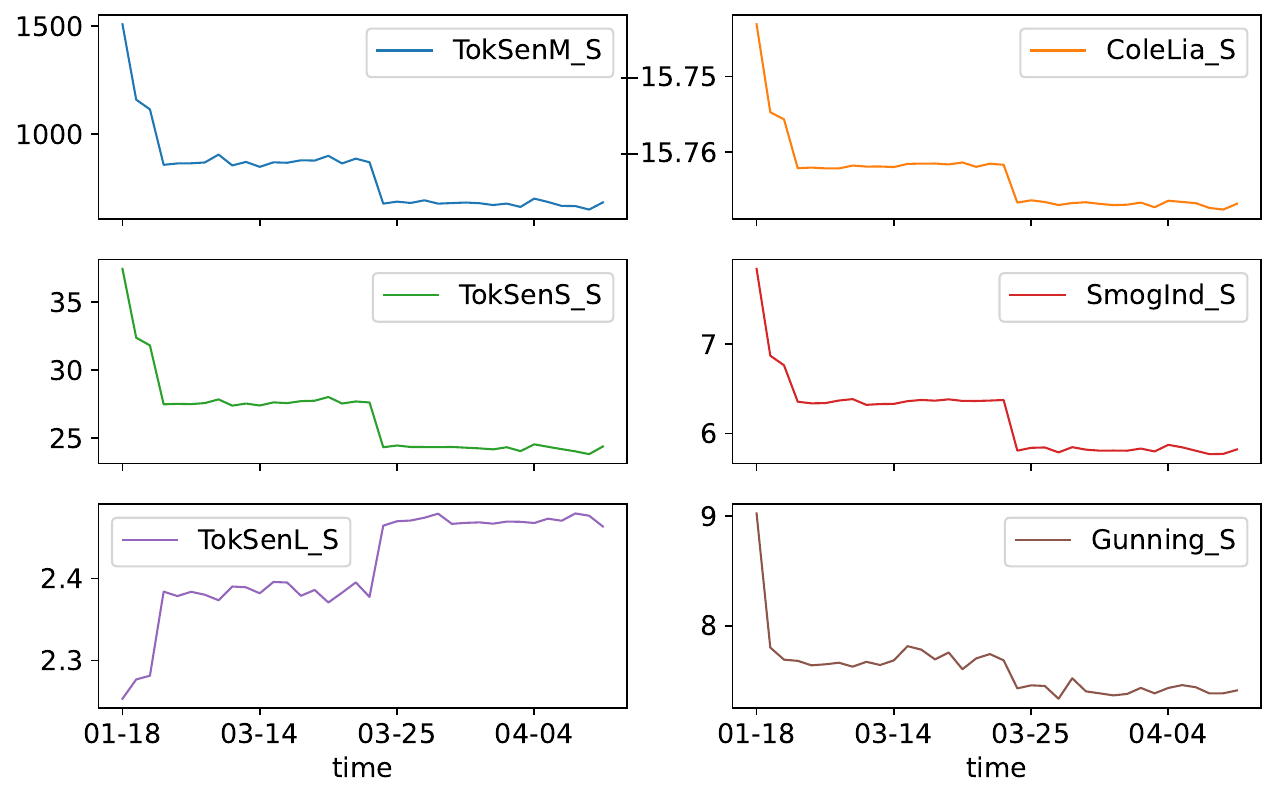}
    \caption{ 
   Trend of ChatGPT's Syntactic (ShaF) linguistic features including total count of tokens (Tok\_S) and readability scores on ChatLog-Daily dataset every day from March 5 to April 9, 2023. Note that the data at start point of January 18 comes from HC3 dataset. 
    }
    \label{fig:ShaF_v1}
\end{figure}

\paragraph{Further Trend Analysis} We show the linguistic features' trends from Figure ~\ref{fig:Adsem_v1} to Figure ~\ref{fig:ShaF_v1}. There are 5 feature categories including Advanced Semantic (AdSem), Discourse (Disco), Syntactic (Synta), Lexico Semantic (LxSem) and Syntactic (ShaF). 

One interesting trend observed in the Advanced Semantic feature category from Figure ~\ref{fig:Adsem_v1} is the increase in the mean value of Clarity (BClar) features over time. This suggests that ChatGPT has become more effective in making its expression more clear in responses over time. Similarly, the number of topic (BTopc20\_S) features decreases across time, which  suggests that ChatGPT has become more accurate in generating fluent responses without loading its writing with fancy topics.

In the Discourse category from Figure ~\ref{fig:Disco_v1}, we observed a simultaneous drop in the number of entities (Enti\_C) and Local Coherence from Entity Grid (LoCoh\_S) scores, indicating that ChatGPT is utilizing more concrete and clear words rather than specific entities in its responses over time, which makes itself easier understood.

In the Syntactic feature category from Figure ~\ref{fig:Synta_v1}, we observed a decrease in the length of Flattened parsed Trees (FTree\_C) and Local Coherence from count of Noun Tags (NoTag\_C) scores, which indicates that ChatGPT has become more effective in structuring sentences so that the distance between dependent words in the flattened parsed trees is minimized. This trend suggests that ChatGPT is generating more coherent and organized sentences over time.

In terms of the Lexico Semantic (LxSem) feature category from Figure ~\ref{fig:LxSem_v1} and ~\ref{fig:LxSem_v2}, we observed a decrease in the number of Noun Variation (NOV\_S) and the Age of Acquisition of words by Kuperman (AAKuW\_C), indicating that ChatGPT is utilizing more frequently occurring words in its responses over time. This trend is consistent with an increasing ability to generate focused and accurate responses.

Finally, in the ShaF feature category from Figure ~\ref{fig:ShaF_v1}, we observed that the total count of tokens (Tok\_S) and readability scores decreased over time. This suggests that ChatGPT has become more concise in generating responses with less words, resulting in a more precise ability to convey meaning.

Taken as a whole, these findings suggest that ChatGPT has been developed and improved over time with an increasing ability to generate coherent, fluent, and concise responses. Analysis of linguistic features provides insights into the specific aspects of understanding  ChatGPT's evolving patterns across time.


\newpage
\paragraph{Further Correlation Analysis}

\begin{figure*}[htbp] 
\centering 
\includegraphics[width=\linewidth]{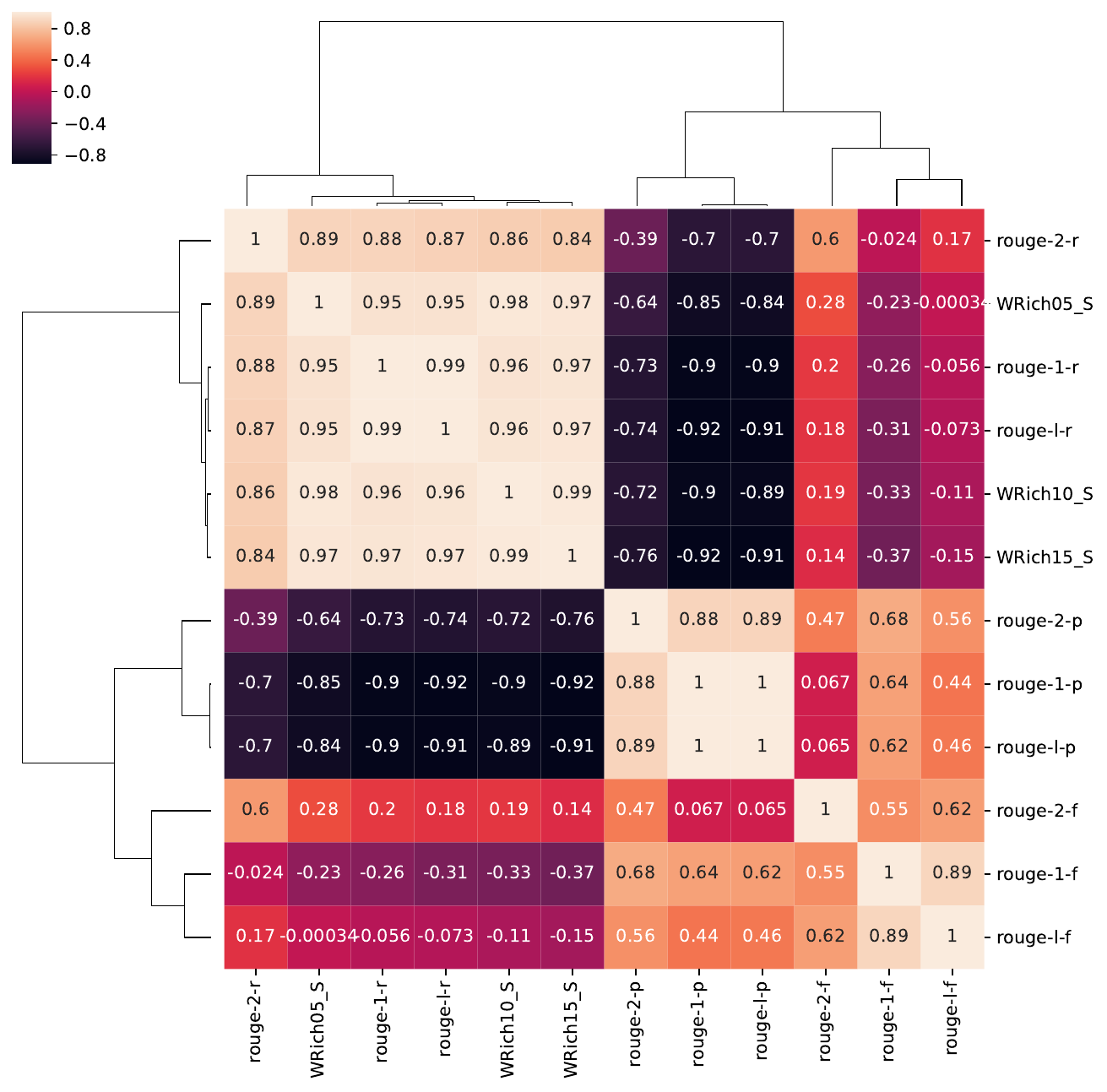} 
\caption{\textbf{Pearson correlation coefficient} between the
nine rouge metrics rouge-\{1,2,l\}-\{precision, recall, f1\} and  three semantic features  WRich\{05,10,15\}\_S, on the ChatLog-Daily dataset after normalization. WRich\_S denotes Semantic Richness coverage on several topics extracted from Wikipedia.} 
\label{fig:full_clustermap} 
\end{figure*}

To comprehensively verify the correlation between extracted features and performance scores, we calculate the Pearson correlation coefficient~\cite{cohen2009pearson} between the
nine rouge metrics rouge-\{1,2,l\}-\{precision, recall, f1\} and  three semantic features  WRich\{05,10,15\}\_S. The normalized results of ChatLog-Daily are shown in Figure \ref{fig:full_clustermap}. We can see that the WRich\_S semantic richness feature scores positively correlate with the rouge-\{1,2,l\}-r performance scores, which can be concluded from the correlation scores in the upper left part of the matrix.

\paragraph{Further Variation Analysis}
In Table \ref{tab:bottom10_0118_0409}.  we calculate the variation coefficient on ChatLog-Daily with data from March 5, 2023, to June 10, 2023, and show features with minimum variation coefficient. These stable features may not change significantly over time, which can be used for downstream tasks. However, if we change the time span of the ChatLog-Daily, will these selected features stay minimum variation on new data? To answer this question, we calculate the variation coefficient on ChatLog-Daily with data from March 5, 2023, to March 28, 2023, and show features with minimum variation coefficient in Table \ref{tab:bottom10_0118_0328}. In fact, the 10 most stable features on new time spans in Table \ref{tab:bottom10_0118_0328} are identical to the 10 features of longer time span in Table \ref{tab:bottom10_0118_0409}. Only the rank of the ra\_ONToT\_C and at\_VeTag\_C feature exchanges with each other while this doesn't affect the set of the 10 most stable features. This phenomenon shows that even after changing the time span, our discovered features still have the minimum variation coefficient.

\begin{table*}[t]
    \centering
    \begin{tabular}{c c c| c c c}
        \toprule
       \textbf{Branch} & \textbf{Feature} & \textbf{Feature Description} & $|\mu|$ & $|\sigma|$ & $\frac{|\sigma|}{|\mu|}$ \\
        \midrule 
       ShaTr & ColeLia\_S & Coleman Liau Readability Score & 15.76 & 1.7e-4 & 1.1e-5 \\
    Disco & ra\_NNTo\_C & Ratio of nn transitions to total & 0.97 & 2.1e-4 & 2.2e-4 \\
   AdSem &  BClar20\_S & Semantic Clarity from 200 topics & 0.80 & 7.9e-4 & 9.9e-4 \\
   AdSem & BClar15\_S &Semantic Clarity from 150 topics & 0.77 & 9.8e-4 & 1.3e-3 \\
   LxSem & BiLoTTR\_S & Type Token Ratio Features & 0.82 & 1.8e-3 & 2.2e-3 \\
   Synta & at\_FTree\_C  & Average length of flattened Trees per token & 1.24 & 2.9e-3 & 2.4e-3 \\
   Synta & at\_ContW\_C & Average count of Content words per token & 0.64 & 1.9e-3 &  2.9e-3\\
   LxSem & at\_SbL1C\_C & Average SubtlexUS Lg10CD value per token & 3.36 & 1.3e-3 &  3.8e-3\\
   Disco & ra\_ONToT\_C & Ratio of on transitions to total & 2.0e-3 & 9.0e-6 &  4.5e-3\\
      Synta & at\_VeTag\_C & average count of Verb POS tags per token & 0.18 & 8.3e-4 &  4.6e-3\\
        \bottomrule
    \end{tabular}
    \caption{Top 10 stable linguistic features from 2023-01-18 to 2023-03-28. $|\mu|$ , $|\sigma|$ , $\frac{|\sigma|}{|\mu|}$ is respectively the average, variation and variation coefficient of feature scores. }
    \label{tab:bottom10_0118_0328}
\end{table*}

\begin{table*}[t]
    \centering
    \begin{tabular}{c c c| c c c}
        \toprule
       \textbf{Branch} & \textbf{Feature} & \textbf{Feature Description} & $|\mu|$ & $|\sigma|$ & $\frac{|\sigma|}{|\mu|}$ \\
        \midrule 
       ShaTr & ColeLia\_S & Coleman Liau Readability Score & 15.76 & 1.6e-4 & 1.0e-5 \\
    Disco & ra\_NNTo\_C & Ratio of nn transitions to total & 0.97 & 2.5e-4 & 2.5e-4 \\
   AdSem &  BClar20\_S & Semantic Clarity from 200 topics & 0.80 & 8.3e-4 & 1.0e-3 \\
   AdSem & BClar15\_S &Semantic Clarity from 150 topics & 0.77 & 9.8e-4 & 1.3e-3 \\
   LxSem & BiLoTTR\_S & Type Token Ratio Features & 0.82 & 1.9e-3 & 2.3e-3 \\
   Synta & at\_FTree\_C  & Average length of flattened Trees per token & 1.24 & 3.0e-3 & 2.4e-3 \\
   Synta & at\_ContW\_C & Average count of Content words per token & 0.63 & 1.9e-3 &  3.0e-3\\
   LxSem & at\_SbL1C\_C & Average SubtlexUS Lg10CD value per token & 3.36 & 1.3e-3 &  3.8e-3\\
      Synta & at\_VeTag\_C & average count of Verb POS tags per token & 0.18 & 8.6e-4 &  4.8e-3\\
   Disco & ra\_ONToT\_C & Ratio of on transitions to total & 2.2e-3 & 1.1e-5 &  5.2e-3\\
        \bottomrule
    \end{tabular}
    \caption{Top 10 most stable linguistic features from 2023-01-18 to 2023-04-09. $|\mu|$ , $|\sigma|$ , $\frac{|\sigma|}{|\mu|}$ is respectively the average, variation and variation coefficient of feature scores. }
    \label{tab:bottom10_0118_0409}
\end{table*}

\clearpage
\section{Details of Detectors}
\label{appendix:details_of_detect}
In this section, we first introduce the details of our training and evaluation of the ChatGPT detection task. Then we will show the baselines' settings, hyper-parameters, and the full detection results on the everyday corpus of ChatLog-Daily. 
\begin{table}[ht]

    \centering
    \begin{tabular}{c|c|c}
        \toprule
\textbf{Module}   &   \textbf{Parameter} & \textbf{Value}   \\
                \midrule 
         
  \multirow{3}{*}{ \shortstack{Feature \\Extraction} }  
  & $n$   & 1000   \\
  & $k$  & 98   \\
     &  $m$   & 265 \\
         \midrule 

 \multirow{8}{*}{ \shortstack{Application\\ of   \\LightGBM} } 
  &boosting type  & gdbt  \\
  & learning rate  & 0.05   \\
  & num\_leaves  & 31   \\
        & feature fraction  & 0.9  \\
    & bagging fraction  & 0.8   \\
   & bagging freq   & 5  \\
      & verbose   & 0  \\
         & boost round   & 50  \\
            & early stop rounds   & 10  \\
        \bottomrule
    \end{tabular}
               \caption{Hyper-parameters for each module.  }
    \label{tab:hyper_parameters}
\end{table}

\subsection{Implementation Details}
\label{appendix:implementation_details}
In our study, we utilized a binary classification approach to investigate whether they can detect ChatGPT responses over time, to maintain robustness on potential changes in ChatGPT's new versions. To do so, we designed a detector based on the RoBERTa~\cite{liu2019roberta} pre-trained language model, which has demonstrated outstanding performance in various NLP tasks.

Our detector baseline is trained on a sample of HC3~\cite{guo2023close} dataset comprising human responses and ChatGPT responses, which are used to fine-tune the RoBERTa model. During fine-tuning, the RoBERTa model is trained to predict whether a given response is human or ChatGPT given question and answer (RoBERTa-qa\footnote{\url{https://huggingface.co/spaces/Hello-SimpleAI/chatgpt-detector}}) or only answer (RoBERTa-single\footnote{\url{https://huggingface.co/spaces/Hello-SimpleAI/chatgpt-detector-single}}). Specifically, we directly use the trained checkpoints from huggingface.

While the RoBERTa pre-trained language model has been shown to be particularly effective in various NLP applications, it should be noted that a major disadvantage of using a pre-trained model as a detector is the lack of interpretability. That is, the model is essentially a black box, making it difficult to understand the specific features that it uses to classify responses. To solve this potential problem and enhance the robustness of RoBERTa, we use an ensemble model LightGBM~\cite{ke2017lightgbm} to train a logistic regression model on the output probability of RoBERTa-single and extra 10 stable features discovered during feature extraction.

The hyper-parameters of our experiment's each module are summarized in Table \ref{tab:hyper_parameters}. For the RoBERTa-based baselines, we run their inferance process on each response of our ChatLog-Daily dataset to collect the predicted classification probability using a single NVIDIA GeForce RTX 3090 GPU. Then the probability is combined with linguistic features to train and test the LightGBM regression model using CPU.



\subsection{Full Detection Results}
\label{appendix:idetection_results}
In fact, the test dataset in Section \ref{application} is sampled from ChatLog-Daily and is blended with human responses to make the label balanced. Here we will show all the classification results for each day's full corpus without balanced human response labels.

\begin{figure*}[htbp] 
\centering 
\includegraphics[width=\linewidth]{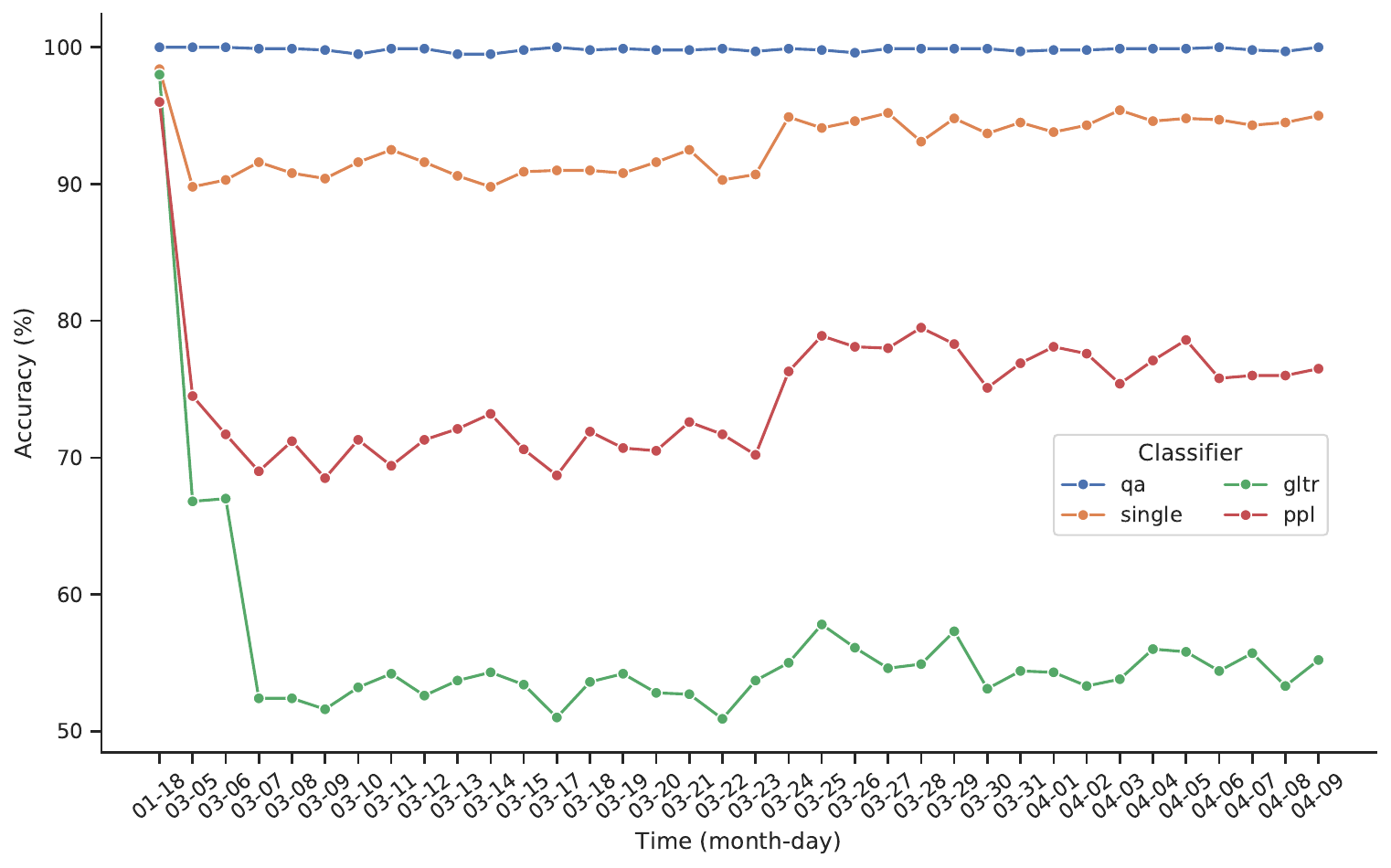} 
\caption{Accuracy of classifiers trained on HC3-01-18 corpus to predict the same 1000 questions' answers collected during full following days in March and April, where each day's data are unbalanced 1000 ChatGPT responses without human responses.} 
\label{fig:full_avg_HC3_classify_line_acc} 
\end{figure*}

\begin{figure*}[htbp] 
\centering 
\includegraphics[width=\linewidth]{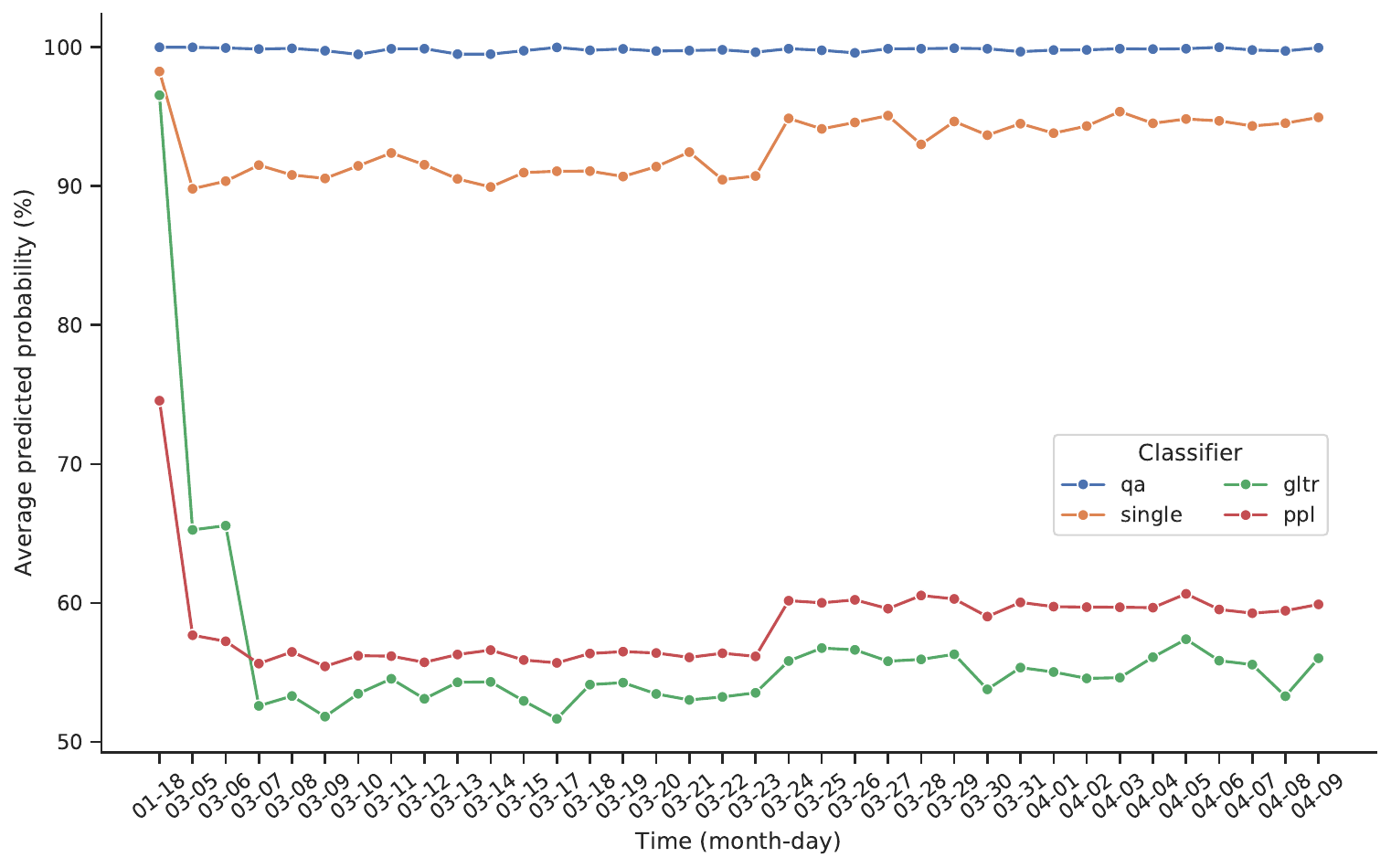} 
\caption{Output ChatGPT classification probability (confidence) of classifiers trained on HC3-01-18 corpus to predict the same 1000 questions' answers collected during the following days in March and April, where each day's data are unbalanced 1000 ChatGPT responses without human responses.} 
\label{fig:avg_HC3_classify_line_prob} 
\end{figure*}

 With classification results in Figure ~\ref{fig:full_avg_HC3_classify_line_acc} and Figure ~\ref{fig:avg_HC3_classify_line_prob}, we have following observations:

(1)  The accuracy of RoBERTa-single, gltr and ppl classifiers  decreases rapidly from 2023-01-18 corpus of HC3 dataset to newly collected corpus in March and April of ChatLog-Daily, which shows the difference between ordinary ChatGPT texts in January and up-to-date texts in March and April. Besides, the texts' classification accuracy also has continuous marginally fluctuations during March and April, which demonstrates that there may exist timely updating of ChatGPT as mentioned by OpenAI.

(2)  Notice that the RoBERTa-qa classifier encodes both question and answer to classify whether the answer is written by ChatGPT. As our everyday asking questions stay the same with the training set, it may cheat on the question information to get high accuracy. If we remove the question information (i.e. RoBERTa-single), its performance drops to 90\%.

(3) Among classifiers, gltr and ppl depend on linguistic features, which have the larger performance drop than fine-tuned classifiers based on RoBERTa, which demonstrates the robustness of the pre-trained language model based classifiers.

\end{document}